\documentclass[review]{elsarticle}

\usepackage{natbib}
\usepackage{csquotes}

\usepackage{mathtools}
\usepackage{amssymb}
\usepackage{amsmath}
\usepackage{algorithm}
\usepackage[noend]{algpseudocode}
\usepackage{lineno,hyperref}
\modulolinenumbers[5]

\journal{Future Generation Computer Systems}

\newcommand{\norm}[1]{\left\lVert#1\right\rVert}

\begin{document}

\begin{frontmatter}

\title{GANterfactual - Counterfactual Explanations for Medical Non-Experts using Generative Adversarial Learning 
}

\author[1]{Silvan Mertes}
\author{Tobias Huber}
\author{Katharina Weitz}
\author{Alexander Heimerl}
\author{Elisabeth Andr\'e}
\address{Augsburg University, Lab for Human-Centered Artificial Intelligence}
\address{Universitätsstraße 6a, Augsburg, Germany}
\address{firstname.lastname@informatik.uni-augsburg.de}

\fntext[fn1]{Corresponding author}

\begin{abstract}
With the ongoing rise of machine learning, the need for methods for explaining decisions made by artificial intelligence systems is becoming a more and more important topic.
Especially for image classification tasks, many state-of-the-art tools to explain such classifiers rely on visual highlighting of important areas of the input data. Contrary, counterfactual explanation systems try to enable a counterfactual reasoning by modifying the input image in a way such that the classifier would have made a different prediction.
By doing so, the users of counterfactual explanation systems are equipped with a completely different kind of explanatory information. However, methods for generating realistic counterfactual explanations for image classifiers are still rare. Especially in medical contexts, where relevant information often consists of textural and structural information, high-quality counterfactual images have the potential to give meaningful insights into decision processes.
In this work, we present \emph{GANterfactual}, an approach to generate such counterfactual image explanations based on adversarial image-to-image translation techniques. 
Additionally, we conduct a user study to evaluate our approach in an exemplary medical use case.
Our results show that, in the chosen medical use-case, counterfactual explanations lead to significantly better results regarding mental models, explanation satisfaction, trust, emotions, and self-efficacy than two state-of-the art systems that work with saliency maps, namely LIME and LRP.
\end{abstract}

\begin{keyword}
generative adversarial networks, explainable AI, machine learning, counterfactual explanations, interpretable machine learning, image-to-image translation
\end{keyword}

\end{frontmatter}

%\linenumbers

\section{Introduction}\label{sec:introduction}
With the rapid development of machine learning (ML) methods, black-box models powered by complex ML algorithms are increasingly making their way into  high risk applications, such as healthcare \cite{stone2016artificial}.
Systems used here must provide comprehensible and transparent information about their decisions.
Especially for patients, who are mostly no healthcare experts, comprehensible information is extremely important to understand diagnoses and treatment options (e.g., \cite{zucco2018explainable, holzinger2017we}).
To support more transparent Artificial Intelligence (AI) applications, approaches for Explainable Artificial Intelligence (XAI) are an ongoing topic of high interest \cite{arrieta2020explainable}.
Especially in the field of computer vision, a common strategy to achieve this kind of transparency is the creation of saliency maps that highlight areas in the input that were important for the decision of the AI system.
The problem with those explanation strategies is, that they require the user of the XAI system to perform an additional thought process: 
Having the information \textit{what} areas were important for a certain decision inevitably leads to the question \textit{why} these areas were of importance. 
Especially in scenarios where relevant differences in the input data are originating from textural information rather than spatial information of certain objects, it becomes clear that the raw information about where important areas are is not always sufficient. In medical contexts, this textural information is often particularly relevant, as abnormalities in the human body often reveal as deviations in tissue structure.
One XAI approach that goes another way to avoid the aforementioned problems of not taking textural information into account are \textit{Counterfactual Explanations}.

Counterfactual explanations try to help to understand why the actual decision was made instead of another one, by creating a slightly modified version of the input which results in another decision of the AI \cite{wachter2017counterfactual,byrne2019counterfactuals}.
Creating such a slightly modified input that changes the model's prediction is by no means a trivial task.
Current counterfactual explanations often utilize images from the training data as basis for modified input images. 
This often leads to counterfactual images that are either distinct but similar images from the training data, or that are unrealistically modified versions of the input image.
Humans, however, prefer counterfactuals that modify as little as necessary and are rooted in reality \cite{byrne2019counterfactuals}.
In this work we present a novel model-agnostic counterfactual explanation approach that aims to tackle these current challenges by utilizing adversarial image-to-image translation techniques.
Traditional generative adversarial networks for image-to-image translation do not take the model into account and are therefore not suited for counterfactual generation.
To this end, we propose to include the classifier's decision into the objective function of the generative networks. 
This allows for the creation of counterfactual explanations in a highly automated way, without the need for heavy engineering when adapting the system to different use cases.

We evaluate our approach by a computational evaluation and a user study inspired by a healthcare scenario. 
Specifically, we use our system to create counterfactual explanations for a classifier that was trained on a classification task to predict if x-ray images of the human upper body are showing lungs that are suffering from pneumonia or not.
In addition to being a highly relevant application for explanations, this scenario is suitable for evaluating explanations for non-experts since they are not expected to have in-depth knowledge of that domain, i.e. they are completely reliant on the explanation that the XAI system gives in order to follow the AI's decisions.
Furthermore, pneumonia in x-ray images predominantly is reflected by opacity in the shown lungs.
Opacity is a textural information that can not be explained sufficiently enough by the spatial information provided by common saliency map approaches.
To validate our assumptions, we compare the performance of our approach against two established saliency map methods,
namely \textit{Local Interpretable Model-agnostic Explanations} (LIME) and \textit{Layer-wise Relevance Propagation} (LRP).\\
With our work we make the following contributions:
\begin{itemize}
    \item We present a novel approach for generating counterfactual explanations for image classifiers and evaluate it computationally.
    \item We evaluate our approach in a user study and gain insights in the applicability of counterfactual explanations for non-ML experts in an exemplary medical context.
    \item We compare counterfactual explanations against two state-of-the-art explanation systems that use saliency maps.
\end{itemize}
The remainder of this work is structured as follows:\\
In Section \ref{sec:related_work}, we give an overview of related topics of XAI. 
Section \ref{sec:approach} introduces our approach in detail, while Section \ref{sec:implementation} presents implementation details and the potential use-case of explaining a classifier in the context of pneumonia detection.
We describe our user study in Section \ref{sec:study}, before we reveal our results in Section \ref{sec:results}. 
We discuss them in Section \ref{sec:discussion}, before we draw conclusions and give an outlook on future research topics in Section \ref{sec:conclusion}.

\section{Related Work}\label{sec:related_work}
\subsection{Explainable AI}
Explainable AI aims to make complex machine learning models more transparent and comprehensible for users. 
To this day, different XAI approaches have emerged that can primarily be distinguished between \textit{model-agnostic} and \textit{model-specific} techniques.
Model-agnostic interpretation methods are characterized by the fact that they are able to provide explanations independent of the underlying model type \cite{molnar2019:InterpretableMachineLearning}.
Model-specific approaches on the other hand exploit the underlying inherent structures of the model and its learning mechanism. As a result they are bound to one specific type of model \cite{molnar2019:InterpretableMachineLearning,Rai2020:blackboxtoglassbox}.
However, even though model-agnostic approaches can easily be applied to a variety of machine learning models they often rely on approximation methods, which in return may impair the quality of explanations, whereas model-specific approaches, due to being specialized on a certain type of machine learning model, usually provide more accurate explanations \cite{hall2018introductionInterpretability}.
A state-of-the-art representative for a model-agnostic approach is LIME \cite{ribeiro2016:lime}. The basic idea of LIME is to approximate an interpretable model around the original model. As a consequence it is possible to create explanations for various machine learning domains like text and image classification.
Depending on the model to be explained the explanations come in the form of textual or visual feedback. In the case of image classification, LIME is highlighting the areas in the image that have been crucial for the prediction of a specific class. LIME has been applied to various healthcare applications like the automatic detection of COVID-19 based on CT scans and chest x-ray images \cite{ahsan2020:xrayLime}, the prediction of a patient's pain \cite{weitz2019deep}, or the prediction of a patient's heart failure risk by utilizing Recurrent Neural Networks \cite{Khedkar2020:LimeHeartFailure}.\\
Lapushkin et al. \cite{bach2015,lapuschkin2019} introduced LRP, a model-specific approach that assigns a relevance value to each neuron in a neural network, measuring how relevant this neuron was for a particular prediction.
For this assignment, they defined different rules, all of which are based on the intermediate outputs of the neural network during the forward pass.
One of those rules introduced by Huber et al. \cite{huber2019} tries to create more selective saliency maps by only propagating the relevance to the neuron with the highest activation in the preceding layer. LRP has been used in different healthcare applications, e.g., analysis of EEG data with deep neural networks \cite{STURM2016:LRPEEG}, histopathological analysis \cite{hagele2020:LRPTumor} and neuroimaging data analysis \cite{Thomas:LRPNeuro} with deep learning.

Besides such feature importance approaches (often called saliency maps in the image domain), that try to identify which features have been most important for predicting an outcome, there are algorithms available that try to answer the question ``How would I have to change my input so that I get a different outcome?''. Those type of explanations are called counterfactual explanations. In fact, counterfactual explanations describe an alternative reality that is contrastive towards the observed one \cite{molnar2019:InterpretableMachineLearning}. This approach of generating explanations is in line with how humans explain things. Humans rarely ask why something happened, but rather why the current outcome is present instead of a different one \cite{miller2019explanation}. This similarity is one of the advantages over approaches that focus on feature importance. 
%Due to their distinct traits, counterfactual explanations have been applied to various domains \cite{mothilal2020explaining, poyiadzi2020face, van2019interpretable}.
Various approaches to generate counterfactual explanations have emerged. The first to introduce counterfactual explanations have been Wachter et al. \cite{wachter2017counterfactual}. They formulated the computation of counterfactuals as an optimization problem. Their goal is to identify a counterfactual that is the closest to the original input, by minimizing the distance between the input data and a potential counterfactual.
Van Looveren et al. \cite{van2019interpretable} propose a model-agnostic approach to generate counterfactual explanations by using class prototypes to improve the search for interpretable counterfactuals. They evaluated their approach on the MNIST dataset, as well as the Breast Cancer Wisconsin(Diagnostic) dataset.
Goyal et al. \cite{goyal2019counterfactual} present an approach to create counterfactual explanations for an image classification task. They exchange a patch of the original image with a patch from a similar image from the training dataset which gets classified differently. They evaluated their approach on four different datasets, including MNIST, SHAPES, Omniglot and Caltech-UCSD Birds (CUB) 2011.

\subsection{Adversarial Approaches to Counterfactual Image Generation}
The first Generative Adversarial Networks (GANs) \cite{goodfellow2014generative} transformed random noise vectors to new image data.
Olson et al. \cite{olson2019counterfactual} use a combination of such a GAN and a Wasserstein Autoencoder to create counterfactual states to explain Deep Reinforcement Learning algorithms for Atari games. 
Nemirovsky et al. \cite{nemirovsky2020countergan} proposed \textit{CounterGAN}, an architecture in which a generator learns to produce residuals that result in counterfactual images when added to an input image.
More recent GAN architectures allow the transformation of images between different image domains \cite{isola2017image,zhu2017unpaired}, known as \textit{Image-to-Image Translation}.
There is existing work that uses those techniques of adversarial image-to-image translation for creating counterfactuals, but often the counterfactuals are not created for the purpose of explaining ML algorithms, but rather to improve those algorithms \cite{neal2018open, wang2020bilateral}. 
Zhao et al. \cite{zhao2020fast} proposed to use a StarGAN \cite{choi2018stargan} architecture to create counterfactual explanation images. 
However, the system was only applied on binary images, i.e. images where each pixel is either black or white. 
The resulting counterfactuals were used to highlight the pixels which differ between original and counterfactual images.

\section{Approach}\label{sec:approach}
In the following sections, we present a novel approach for generating counterfactual explanations for image classifiers using generative adversarial learning.

\subsection{Counterfactual Explanations as an Image-to-Image Translation Problem}
As discussed by Wachter et al. \cite{wachter2017counterfactual}, one of the key concepts of counterfactual explanations is the concept of the \textit{closest possible world}. 
Counterfactual explanations aim to show a slight variation of some object, where the change between the original object and its variation results in a different outcome. 
Transferred to the task of explaining image classifiers, counterfactual explanations should aim to answer the following question:
 \begin{quote}
    \textit{What minimal changes to the input image would lead the classifier to make another decision?}
 \end{quote}
This question implicates two major requirements to counterfactual images: 
\begin{itemize}
	\item The counterfactual image should look as similar to the original image as possible.
	\item The classifier should predict the counterfactual image as belonging to another class as the original image.
\end{itemize}
Looking at the second statement at a more abstract level, the predicted class of an image can be seen as some sort of top-level feature that describes a combination of several underlying features which the classifier considers to be relevant for the classification. 
Thus, the generation of counterfactual images can be broken down to a transformation of certain features that are relevant for the classification, 
while maintaining all other features, which were not relevant for the classification.
However, these two objectives are also defining the problem of \textit{Image-to-Image Translation}.
The goal of image-to-image translation is to transform features that are relevant for a certain image domain to features that lead to another image domain, while all other features have to be maintained.
An example of such an image-to-image translation task are style-conversion problems, where each image domain represents a certain style.
In this case, translating an image from one domain to another is equivalent to changing the style of the image.
Viewing the problem of counterfactual creation from the perspective of image-to-image translation inevitably leads to the idea of borrowing techniques from that area for generating counterfactual images to explain image classifiers.   

\subsection{Image-to-Image Translation with CycleGANs}
There are various approaches for solving image-to-image translation problems.
Recent promising approaches rely on the use of adversarial learning.
The original GANs \cite{goodfellow2014generative} 
approximate a function that transforms random noise vectors to images which follow the same probability distribution as a training dataset (i.e., that appear similar to images from the training set which the GAN was trained on).
They do this by combining a \textit{generator network $G$} and a \textit{discriminator network $D$}.
During training the generator learns to create new images, while the discriminator learns to distinguish between images from the training set and images that were created by the generator. 
Thus, the two networks are improving each other in an adversarial manner.
The objective of the two networks can be defined as follows:
\begin{multline}
\mathcal{L}_{original}(G, D)=
\mathbb{E}_{x\sim p_{data} (x)}[\text{log } D(x)] + \mathbb{E}_{z\sim p_{z} (z)}[\text{log } (1 - D(G(z)))],
\end{multline}
where $x$ are instances of image-like structures and $z$ are random noise vectors.
During training, the discriminator $D$ maximizes that objective function, while the generator $G$ tries to minimize it.

Various modified architectures have successfully been used to replace the random input noise vectors with images from another domain.
Thus, those architectures are capable of transforming images from one domain to images of another domain. 
These approaches are commonly described as image-to-image translation networks. 
Common adversarial approaches for these kind of tasks rely on paired datasets (i.e., datasets that consist of pairs of images which only differ in the features that define the difference of the two image domains).
As described above, in the context of counterfactual image generation for image classifiers, the aim is to transfer images from the domain of one class to the domain of another class.
The aforementioned adversarial architectures are therefore not suited for the generation of counterfactual images since they could only be applied for classifiers that are trained on paired datasets. 
In practice, paired datasets for image classification are a rare occasion. 
A solution to the problem of paired datasets was posed by Zhu et al. \cite{zhu2017unpaired}, who introduced the \textit{CycleGAN} architecture. This architecture is based on the idea of combining two GANs, where one GAN learns to translate images of a certain domain $X$ to images of another domain $Y$, while the other GAN learns to do the exact opposite: convert images of domain $Y$ to images of domain $X$.
The respective objective is defined as follows:
\begin{multline}
\mathcal{L}_{GAN}(G, D_Y, X, Y)=
\mathbb{E}_{y\sim p_{data} (y)}[log D_Y(y)] + \mathbb{E}_{x\sim p_{data} (x)}[log (1 - D_Y(G(x)))]
\label{eq:adversarial_loss}
\end{multline}
where $G$ is the generator of the first GAN and $D_Y$ the discriminator of the same GAN. Therefore, that first GAN learns the translation from images of domain $X$ to images of domain $Y$. The objective of the second GAN, which consists of a generator $F$ and a discriminator $D_X$, is defined analogously.

By feeding images $x$ of domain $X$ to $G$ and subsequently feeding the resulting image $G(x)$ to $F$, the output of the second GAN $F(G(x))$ can be compared with the initial input $x$ (and vice versa) to formulate a so-called \textit{Cycle-consistency Loss}:
\begin{multline}
\mathcal{L}_{cycle}(G, F)=
\mathbb{E}_{x\sim p_{data} (x)}[\norm{F(G(x)) - x}_1] + \mathbb{E}_{y\sim p_{data} (y)}[\norm{G(F(y)) - y}_1],
\end{multline}
where $\norm{x}_1$ represents the $L1$ norm. 
In combination with the adversarial losses given by Equation \ref{eq:adversarial_loss}, the cycle-consistency loss can be minimized to solve image-to-image translation tasks that do not rely on a dataset of paired images. 
The full objective of such common CycleGANs is denoted as:
\begin{multline}
\mathcal{L}(G, F, D_X, D_Y) =
\mathcal{L}_{GAN}(G, D_Y, X, Y) + \mathcal{L}_{GAN}(F, D_X, Y, X) + \lambda\mathcal{L}_{cycle}(G, F)
\end{multline}
During training, the discriminators $D_X$ and $D_Y$ aim to maximize that objective function, while the generators $G$ and $F$ try to minimize it.

\subsection{Extending CycleGANs for Counterfactual Explanations}

\begin{figure}
 \centering
  \includegraphics[width=1\linewidth]{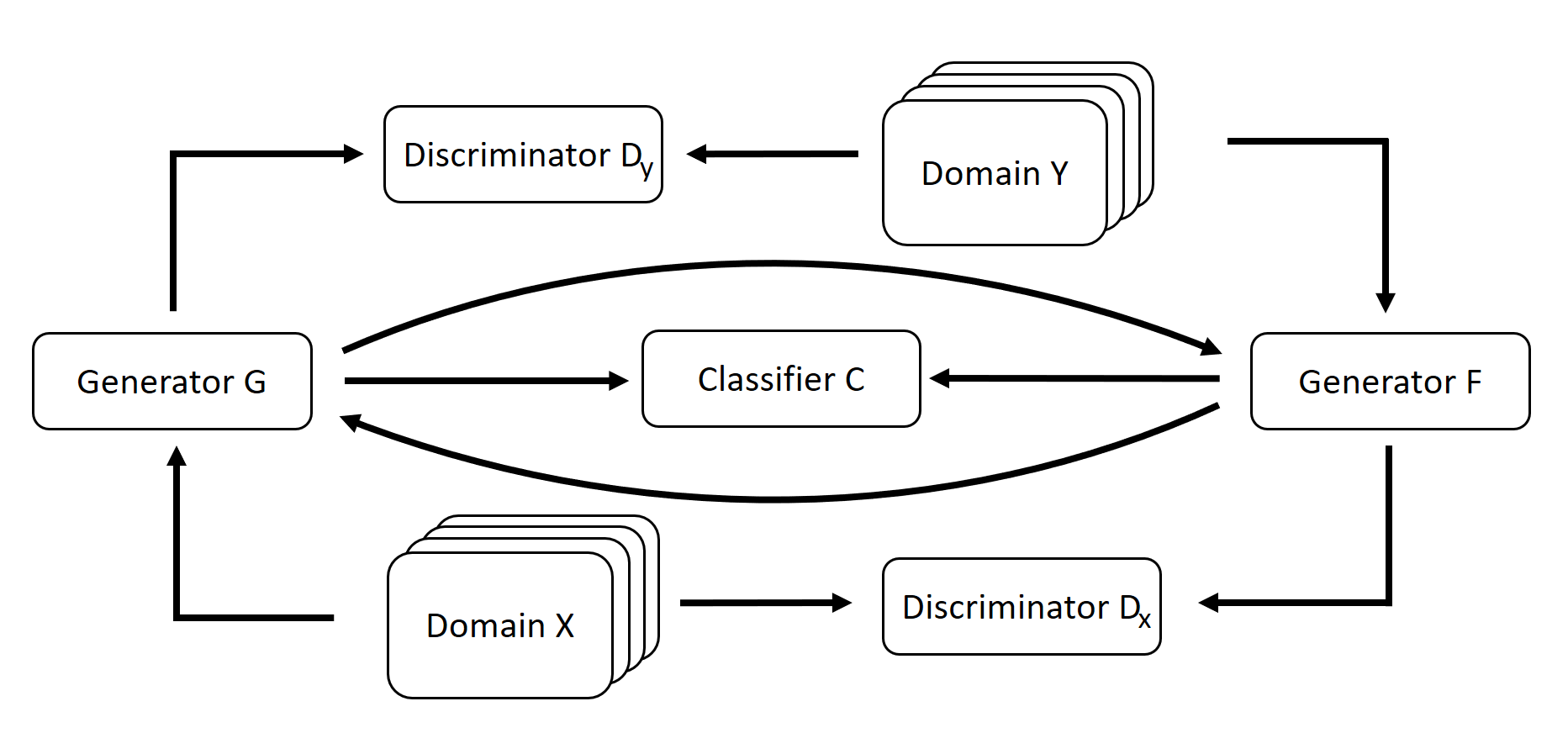}
  \caption{Schematic overview of our approach. A CycleGAN architecture is extended with the classifier that shall be explained. Both the generators of the CycleGAN include the classifier's decisions for the generated data into their loss function.}
  \label{fig:schematic_overview}
\end{figure}

Without loss of generality, we restrict ourselves to the generation of counterfactual explanations for a binary classifier (i.e., a classifier that only decides if an input image belongs to either one class or another).
In theory, this can easily be extended to a multi-class classification problem by looking at each combination of classes as a separate binary problem.
A naive approach to creating counterfactual images for a binary classifier would be to train a traditional CycleGAN architecture to transfer images between the two domains which are formed by the two classes of the training dataset of the classifier.
This would lead to a system that is able to convert images from the domain of one class to images of the domain of the other class, while maintaining features that do not contribute to determining to which domain an image belongs to.
If we now assume that the classifier, which we want to explain, is perfect and always predicts the correct class for every possible image in the two domains, then this would lead to counterfactual explanations: An input image, which was classified to belong to one of the two classes, can be fed into the trained CycleGAN to translate it into an image that is classified as the other class.
However, this might not be an ideal explanation, since it is theoretically possible that the classifier does not use all the features which define a class (e.g., if some features are redundant).
Moreover, the assumption of a perfect classifier is obviously wrong in the most cases.
Thus, the resulting image can by no means be seen as a counterfactual \textit{explanation} of a classifier, as the translation happens between two classes of the training dataset without considering the classifier's decision.
To tackle that problem, a further constraint has to be added to the CycleGAN in order to take the actual decision of the classifier into account.
To achieve this, we propose to incorporate an additional component to the CycleGAN's objective function, which we will describe below.
Analogous to above, where $x$ represented an image of domain $X$, let $x$ now be an image that belongs to class $X$, while $y$ belongs to class $Y$.
Furthermore, consider a classifier $C$ that for every input image $img$ predicts either $C(img) = X$ or $C(img) = Y$. 
In this case, a \textit{perfect} classifier would fulfill both of the following statements:
\begin{align}
    \forall x \in X: C(x) = X  
    \quad \text{and} \quad
    \forall y \in Y: C(y) = Y
\end{align}
As of the objective functions that are used for the definition of the CycleGAN, $G$ is responsible for the translation of images $x$ from domain $X$ to images that belong to $Y$, while $F$ translates images from $Y$ to $X$.
As a counterfactual explanation should show images that the classifier would assign to another class as the original input images, the following statements should be fulfilled by $G$ and $F$ respectively:
\begin{align}
C(img) = X &\implies C(G(img)) = Y \nonumber \\
&\text{\centering and} \nonumber \\
C(img) = Y &\implies C(F(img)) = X 
\end{align}
Most state-of-the-art classifiers do not simply output the actual class that was predicted. They rather use a softmax function to output a separate value for each class, representing the probability that the input actually belongs to the respective class. 
Thus, we extend the above formulation of our binary classifier to $C_2(img) = (p_X, p_Y)^T$, where $p_X$ represents the probability of $img$ belonging to $X$, while $p_Y$ represents the probability of $img$ belonging to $Y$. With this in mind, we can formulate a loss component for the counterfactual generation:
\begin{align}
\mathcal{L}_{counter}(G, F, C) &= 
\mathbb{E}_{x\sim p_{data} (x)}[\norm{C_{2}(G(x)) - \begin{psmallmatrix} 0 \\ 1 \end{psmallmatrix}}_{2}^{2}] \nonumber \\ 
&+ \mathbb{E}_{y\sim p_{data} (y)}[\norm{C_{2}(F(y)) - \begin{psmallmatrix} 1 \\ 0 \end{psmallmatrix}}_{2}^2],
\end{align}
where $\norm{\cdot}_{2}^{2}$ is the squared L2 Norm (i.e., the squared error).

We chose the vector $(1,0)^{T}$ and $(0,1)^{T}$ since we wanted very expressive counterfactuals that are understandable by non-expert users.
In theory one could also chose closer vectors like $(0.49,0.51)$ to enforce counterfactual images that are closer to the decision boundary of the classifier.

Using our proposed counterfactual loss function allows to train a CycleGAN architecture for counterfactual image generation. During training, the generator networks of both GANs are getting punished for creating translated images that are not classified as belonging to the respective counterfactual class by the classifier.\\
Furthermore, as proposed by the authors of CycleGAN \cite{zhu2017unpaired}, we add an \emph{identity loss}, that forces input images to stay the same, if they already belong to the target domain:
\begin{multline}
\mathcal{L}_{identity}(G, F)=
\mathbb{E}_{y\sim p_{data} (y)}[||G(y) - y||_1] + \mathbb{E}_{x\sim p_{data} (x)}[||F(x) - x||_1]
\end{multline}
Thus, the complete objective function of our system is composed as follows:
\begin{align}
    \mathcal{L}(G, F, D_X, D_Y, C) &= \mathcal{L}_{GAN}(G, D_Y, X, Y) \nonumber \\
&+ \mathcal{L}_{GAN}(F, D_X, Y, X) \nonumber \\ 
&+ \lambda\mathcal{L}_{cycle}(G, F) \nonumber \\
&+ \mu\mathcal{L}_{identity}(G, F) \nonumber \\
&+ \gamma\mathcal{L}_{counter}(G, F, C)
\end{align}
where $\mu$ is an \emph{Identity Loss Weight} and $\gamma$ is a \emph{Counterfactual Loss Weight}.
During training, the discriminators $D_X$ and $D_Y$ aim to maximize that objective function, while the generators $G$ and $F$ try to minimize it.\\
A schematic overview of our approach is depicted in Figure \ref{fig:schematic_overview}.

\section{Implementation and Computational Evaluation} 
%\footnote{The code of our implementation can be found at https://github.com/hcmlab/GANterfactual}}
The code of our implementation can be found online.\footnote{\url{https://github.com/hcmlab/GANterfactual}}
\label{sec:implementation}
\subsection{Use Case: Pneumonia Detection}

\begin{figure}
 \centering
  \includegraphics[width=1\linewidth]{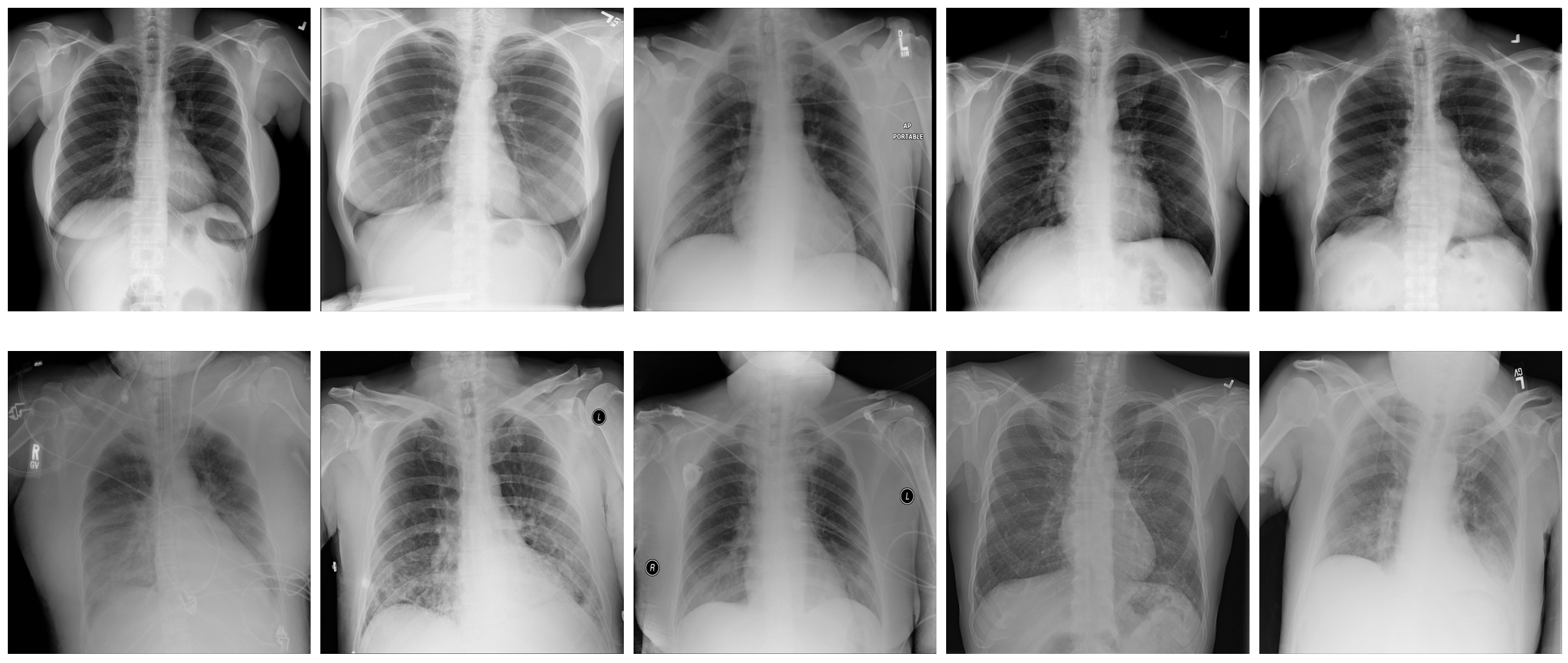}
  \caption{Example images of the used dataset. The top row shows images that are labeled as \textit{Normal}, while the bottom row shows images labeled as \textit{Lung Opacity}, indicating lungs that are suffering from pneumonia.}
  \label{fig:dataset_examples}
\end{figure}

One major drawback of common XAI techniques such as LIME or LRP is, that they highlight certain regions of interest, but they do not tell something about the semantics of that regions. Thus, when explaining a machine learning model, they give information about \textit{where} to look for relevant things, but not explicitly \textit{why} those things are relevant. Counterfactual explanation images tackle this problem. As they alter the original input image, they directly show \textit{how} the input could have looked like, such that another decision would have been made, instead of only showing \textit{where} a modification of the input would make a difference in the classifiers outcome. 
Thus, we argue that the advantages of such a counterfactual system stand out especially in explanation tasks where the users of the system do not have much prior knowledge about the respective problem area and are not able to interpret the semantics of the regions of relevance without assistance. \\
Thus, to evaluate our approach, we chose the problem of \textit{Pneumonia Detection}.
We trained a binary classification Convolutional Neural Network (CNN) to decide whether a given input of a human upper body's x-ray image shows a lung that suffers from pneumonia or not. Subsequently, we trained a CycleGAN that was modified with our proposed counterfactual loss function, incorporating the trained classifier.

We used this problem as a use case for evaluating our system, as medical non-experts do not have a deeply formed mental model of that problem.
We hypothesize that this leads to a lack of interpretability for common XAI techniques that only highlight areas of relevance. 
From an ethical point of view, it is important that medical non-experts understand the diagnoses that relate to them \cite{zucco2018explainable}. 

\subsubsection{Classifier Training}
The aim of this section is to give an overview of the classifier that we want to explain for our particular use case. However, we want to emphasize that our approach is not limited to this classifier's architecture. The only requirement for training our explanation network is a binary classifier $C$ that is able to return a class probability vector $(p_X, p_Y)^T$ for an image that is fed as input.\\
To evaluate our system, we trained a CNN to decide if input images of x-rays are showing lungs that suffer from pneumonia or not. As dataset, we used the data set published for the \textit{RSNA Pneumonia Detection  Challenge}\footnote{https://www.kaggle.com/c/rsna-pneumonia-detection-challenge/} by the  Radiological  Society  of  North America. The original dataset contains 29,700 frontal-view x-ray images of 26,600 patients. The training data is split into three classes: \textit{Normal}, \textit{Lung Opacity} and \textit{No Lung Opacity / Not Normal}. We took only the classes \textit{Normal} and \textit{Lung Opacity}, as Franquet et al. \cite{franquet2018imaging} argue that opacity of lungs is a crucial indicator of lungs suffering from pneumonia, and we only wanted to learn the classifier to distinct between lungs suffering from pneumonia and healthy lungs. Other anomalies that do not result in opacities in the lungs are excluded from the training task to keep it a binary classification problem.  All duplicates from the same patients were removed as well. For the sake of simplicity, we will refer to the class \textit{Lung Opacity} as \textit{Pneumonia} in the rest of this paper. The resolution of the images was reduced to $512x512$ pixels. Subsequently, we randomly split the remaining 14,863 images into three subsets: \textit{train}, \textit{validation}, and \textit{test}.\footnote{The split we used for our experiments is available from the authors upon request.} The distribution of the partitions is shown in Table \ref{table:dataset_distribution}.

\begin{table}[h!]
\centering
\begin{tabular}{||l c c c||} 
 \hline
 Partition & Normal & Pneumonia & Total \\ [0.5ex] 
 \hline\hline
 Train (70\%)& 6195 & 4208 & 10403 \\ 
 Validation (10\%)& 886 & 602 & 1488 \\
 Test (20\%)& 1770 & 1202 & 2972 \\
 \hline
 Total & 8851 & 6012 & 14863 \\ [1ex] 
 \hline
\end{tabular}
\caption{Distribution of the images of the used dataset.}
\label{table:dataset_distribution}
\end{table}

We trained an AlexNET architecture to solve the described task. For details about AlexNET, we want to point the interested reader to Krizhevsky et al. \cite{krizhevsky2017imagenet}. 
We slightly modified the architecture to fit our needs.
These modifications primarily include L2 regularization to avoid overfitting. 
A detailed description of the model that we used can be found in \ref{appendix:classifier_architecture}.
The training configuration is shown in Table \ref{table:classifier_configuration}.

\begin{table}[h!]
\centering
\begin{tabular}{||l c||} 
 \hline
 Parameter & Value  \\ [0.5ex] 
 \hline\hline

 Optimizer & Stochastic Gradient Descent  \\
 Learning Rate & 0.0001 \\ 
 Momentum & 0.9 \\
 Batch Size & 32 \\
 Epochs & 1000 \\
 Loss Function & Mean Squared Error  \\  [1ex] 
 \hline
\end{tabular}
\caption{Training configuration of the used AlexNET.}
\label{table:classifier_configuration}
\end{table}

After training the classifier on the \textit{train} partition for 1000 epochs, it achieved an accuracy of 91,7\% on the \textit{test} set ($f1$ score: 0.894; $f2$ score: 0.883). It should be noted that there exists a plethora of work that focuses on building classifiers that achieve a high classification performance on tasks that are similar to this one.
Those classifiers achieve much better performance values than our classifier does. However, as the aim of our work is to \textit{explain} the decisions of a classifier, we found that a perfect model would not be an appropriate tool to measure the performance of an XAI system. Thus, we did not try to improve the classifier performance further (i.e., we did not conduct any hyperparameter tuning or model optimization). 

\subsubsection{CycleGAN Training}
\begin{figure}
 \centering
  \includegraphics[width=1\linewidth]{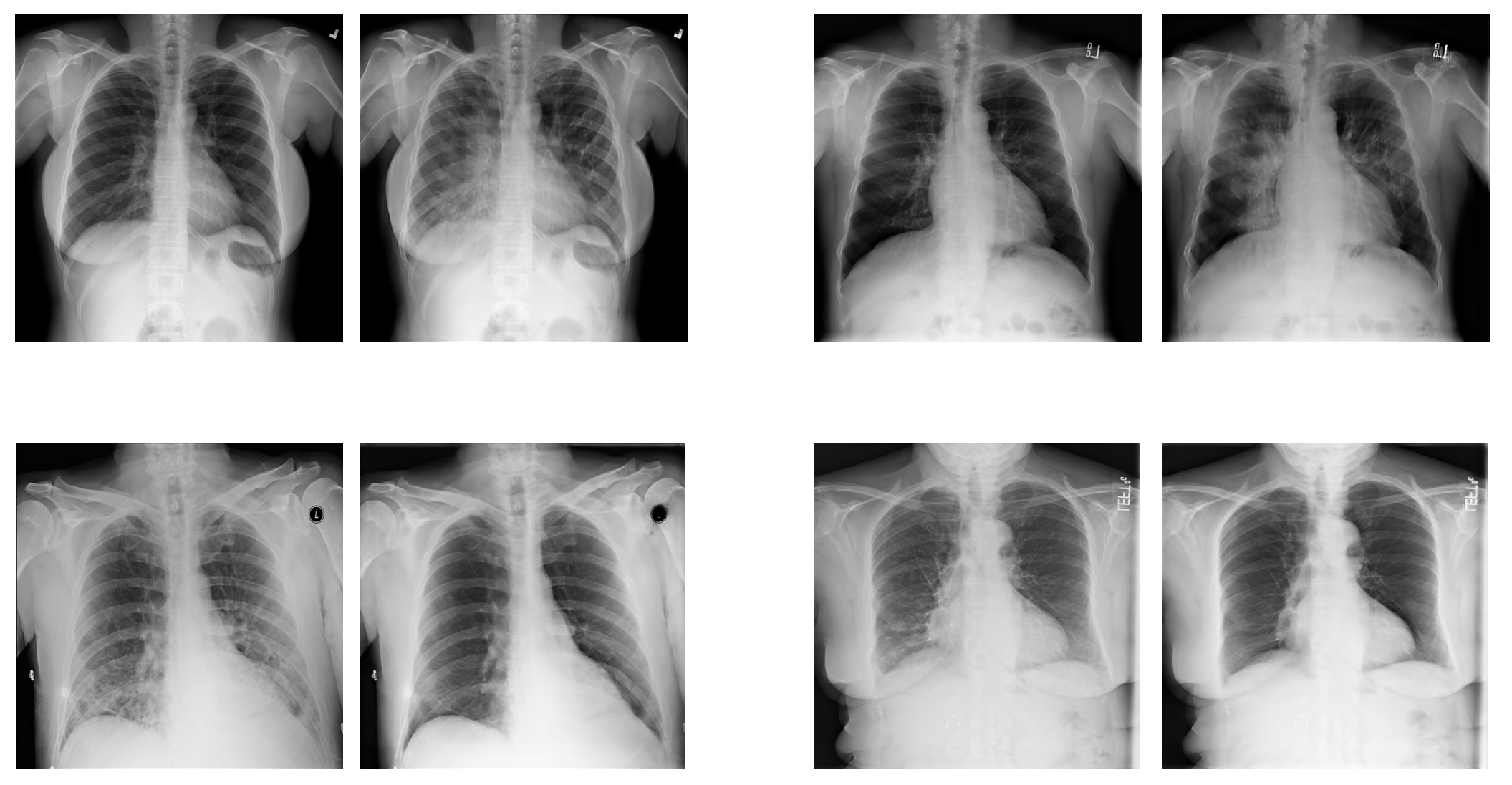}
  \caption{Examples of counterfactual images produced with our proposed approach. In each pair, the left image shows the original image, while the right image shows the corresponding counterfactual explanation. The original images in the top row were classified as \textit{normal}, while the original images in the bottom row were classified as \textit{pneumonia}. The shown counterfactual images were all classified as the opposite as their respective counterpart.}
  \label{fig:counterfactual_examples}
\end{figure}

We trained a CycleGAN model whose objective function was adapted as we propose in Section \ref{sec:approach}. As training dataset, we used the \textit{train} partition of the same dataset that we used for our classifier. Our proposed counterfactual loss $\mathcal{L}_{counter}$ was calculated using the trained classifier that was described in the previous subsection. The architecture of both the generators as well as both the discriminators where adopted from Zhu et al. \cite{zhu2017unpaired}. As proposed by them, we additionaly used a modified version of the discriminator architecture called \textit{PatchGAN}.
This variant of the discriminator approximates validity values for different patches of the input instead of a single validity value for the whole input.
Such a validity value estimates whether the input was generated by the generator or came from the training set.
Further architectural details can be found in their publication. The training configuration parameters are listed in Table \ref{table:cyclegan_configuration}. Examples of counterfactual images that were produced by feeding images from the \textit{test} partition into our trained generative model are shown in Figure \ref{fig:counterfactual_examples}. 
Here, the main structure and appearance of the lungs are maintained during the translation process, while the opacity of the lungs is altered.
This was expected due to the pneumonia class of the used dataset being defined by lungs that show a certain degree of opacity. All in all, the visual inspection of the produced results shows that our approach is promising. \\

\begin{table}[h!]
\centering
\begin{tabular}{||l c||} 
 \hline
 Parameter & Value  \\ [0.5ex] 
 \hline\hline

 Optimizer & Adam  \\
 Learning Rate & 0.0002 \\ 
 Beta 1 & 0.5 \\
 Beta 2 & 0.999 \\
 Batch Size & 1 \\
 Epochs & 20 \\
 Cycle Consistency Loss Weight & 10 \\
 Identity Loss Weight & 1 \\
 Counterfactual Loss Weight  & 1 \\  [1ex] 
 \hline
\end{tabular}
\caption{Training configuration of the CycleGAN with our proposed counterfactual loss function.}
\label{table:cyclegan_configuration}
\end{table}

\subsection{Computational Evaluation}
\begin{figure}[t]
 \centering
  \includegraphics[width=1\linewidth]{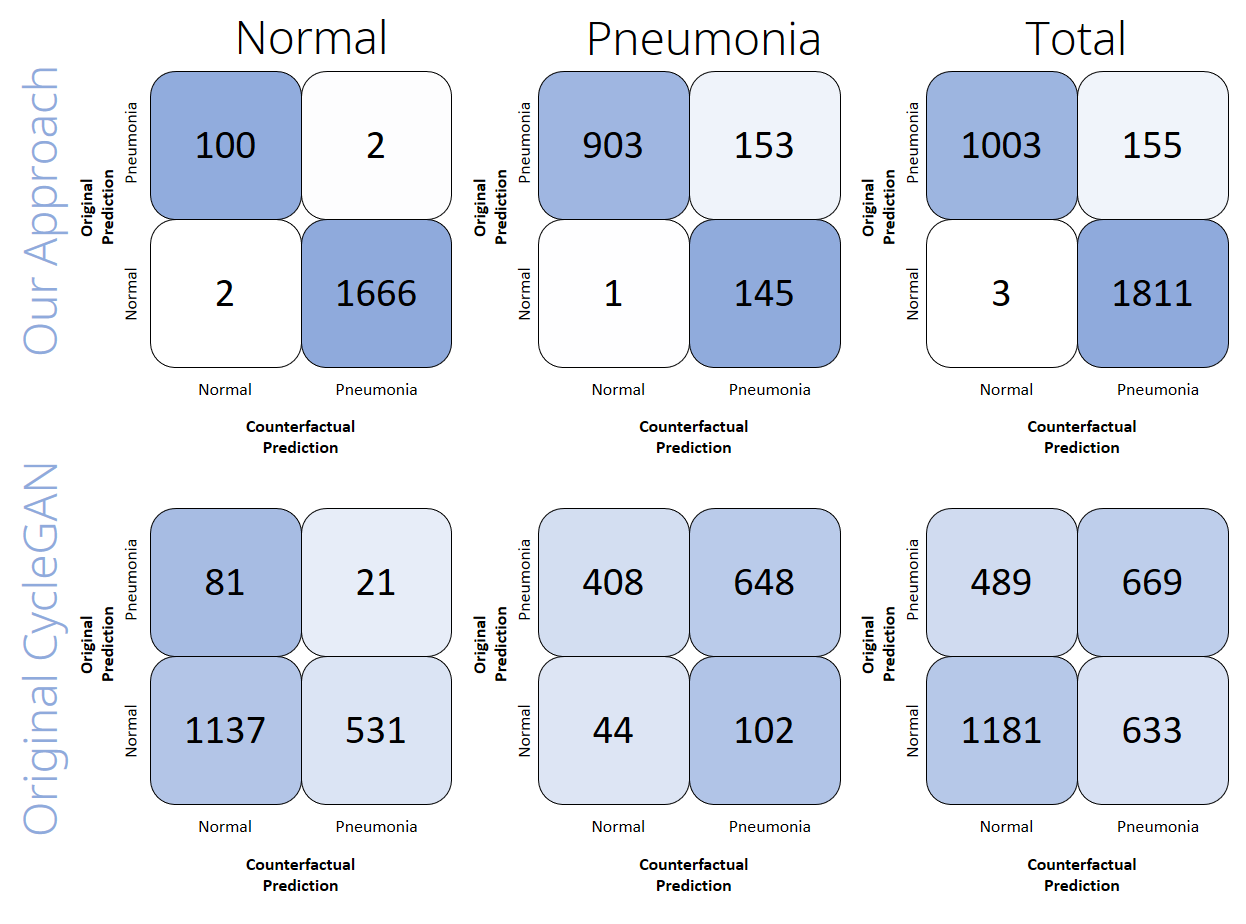}
  \caption{Computational evaluation results of the counterfactual image generation performance. 
  The confusion matrices show the number of samples out of each subset (Normal, Pneumonia, Total) of the rsna dataset that the classifier predicted to be the respective class before (y-axis) and after (x-axis) the samples had been translated by either the original CycleGAN or by our approach.}
  \label{fig:counterfactual_conf_matrix}
\end{figure}

To see if the produced counterfactual images are classified differently than the original input images, we evaluated the system on the \textit{test} partition.
To this end, we fed every image into the classifier, translated the image by the use of the respective generator network, and then classified the resulting counterfactual image. 
We did this separately for the images that originally were labeled as \textit{normal}, as well as for those that were labeled as \textit{Lung Opacity}. We performed this whole procedure for a CycleGAN that was modified with our approach, as well as for an original CycleGAN architecture that does not implement our proposed counterfactual loss function. Figure \ref{fig:counterfactual_conf_matrix} shows the results. 
It can be seen that the counterfactual images generated by our approach were indeed classified as a different class than the original image in most of the cases.
In total, our approach reaches an accuracy of 94.68\%, where we understand the accuracy of a counterfactual image generator to be the percentage of counterfactuals that actually changed the classifier's prediction.
For the images that were originally labeled as \textit{normal}, the accuracy was 99.77\%, while for the images that were labeled as \textit{Lung Opacity} the accuracy reached 87.19\%. 
Contrary, the original CycleGAN only reaches 37.75\% accuracy in total (34.58\% on \textit{normal} lungs, 42.43\% on \textit{Lung Opacity} lungs). 
Those results indicate that the modification of the CycleGAN's objective with our additional counterfactual loss has a huge advantage over the original CycleGANs when aiming for the creation of counterfactual images.
In conclusion, it seems that the counterfactual generation with our approach works sufficiently well, but it has a harder time when confronted with images that actually show lungs suffering from pneumonia than in the case of processing images that show normal lungs. 

\section{User Study}\label{sec:study}
To investigate the advantages and limitations of XAI methods, it is crucial to conduct human user studies.
In this section we describe the user study we conducted to compare our proposed counterfactual approach with two state-of-the-art XAI approaches (LRP and LIME).

\subsection{Conditions}
We compare three independent variables by randomly assigning each participant to one of three conditions.
The participants in each condition only interacted with a single visual explanation method (between-subjects design).
Participants in the LRP condition were assisted by heatmaps generated through Layer-wise Relevance Propagation using the $z$-rule for fully connected layers and the $\alpha 1 \beta 0$-rule for convolutional layers, as recommended by Montavon et al. ~\cite{montavon2019lrp_overview}.
The LIME condition contained highlighted Super-Pixels which were generated by LIME.
Here, we chose the \emph{SLIC} segmentation algorithm which Schallner et al. \cite{schallner2019effect} found to perform well in a similar medical use case. 
For the remaining hyperparameters we used the default values and showed the five most important Super-Pixels.
For both LIME and LRP, we omit the negative importance values since those were highly confusing to participants in our pilot study.
Participants in the counterfactual condition were shown counterfactual images generated by our proposed approach (see section \ref{sec:approach}).
The three different visualisations can be seen in Figure \ref{fig:XAI_vis_study}.

\subsection{Hypotheses}
All our Hypotheses are targeting non-experts in healthcare and artificial intelligence.
Since our aim is to evaluate our proposed counterfactual approach, we do not investigate differences between the saliency map conditions (LRP and LIME).
For our user study we formulated the following hypotheses:
\begin{itemize}
    \item \textbf{Explanation Satisfaction}: Participants are more satisfied with the explanatory quality of counterfactuals compared to LIME and LRP.
    \item \textbf{Mental Models}: Participants use counterfactuals to create more correct mental models about the AI than with LIME and LRP.
    \item \textbf{Trust}: 
    Participants have more trust in the AI system if it is explained with counterfactuals than if it is explained with LRP or LIME.
    \item \textbf{Emotions}: The intuitive and simple interpretation of counterfactuals makes participants feel happier, more relaxed and less angry compared to LRP and LIME.
    \item \textbf{Self-efficacy}: If counterfactuals are a more satisfying XAI method than LRP or LIME, participants feel also strengthened in their self-efficacy towards the AI system, compared to participants in the LRP and LIME conditions.
\end{itemize}

\subsection{Methodology}

To evaluate our hypotheses, we used the following Methods:

\paragraph{Mental Models} 
We use two metrics to evaluate the mental models that the participants formed through our XAI methods. 
Quantitatively, we conduct a prediction task, as proposed by Hoffman et al. \cite{hoffman2018MetricsXAI}, where the participants have to predict what the AI model will decide for a given x-ray image.
For a more qualitative evaluation, we used a form of task reflection, also proposed by Hoffman et al. \cite{hoffman2018MetricsXAI}.
Here, the participants were asked to describe their understanding of the AI's reasoning after they completed the prediction task. 
For this, the participants were asked two questions about their mental model of the AI: ``What do you think the AI pays attention to when it predicts pneumonia?'' and ``What do you think the AI pays attention to when it predicts healthy lungs?''

\paragraph{Explanation Satisfaction} We used the Explanation Satisfaction Scale, proposed by Hoffman et al. \cite{hoffman2018MetricsXAI} to measure the participants' subjective satisfaction with the visual explanations (LRP, LIME, or counterfactuals) that we presented. 

\paragraph{Trust} To evaluate the trust in the presented AI system, we used two items (i.e., ``I trust the system'' and ``I can rely on the AI system'') from the Trust in Automation (TiA) questionnaire proposed by Körber et al. \cite{korber2018theoretical}. Körber points out that one or two items are sufficient to measure trust if people have no previous experience with the system, as is the case with our system.

\paragraph{Emotions} We used items for the subscales \emph{anger}, \emph{happiness}, and \emph{relaxation} of the Discrete Emotions Questionnaire (DEQ)\cite{harmon2016discrete} to evaluate the participants feelings during solving the tasks.

\paragraph{Self-efficacy} We used one item to measure the self-efficacy towards the AI system. For this, we used a variation of one item proposed by Bernacki et al. \cite{bernacki2015examining} (i.e., "How confident are you that you could detect pneumonia using the presented explanations in the future?").

\subsection{Participants}
In order to detect an effect of $\eta_{p}^{2}$ =0.04, with 80 \% power in a one-way between subject MANOVA (three conditions, $\alpha$=.05), the conducted a-priori power analysis suggested that we would need 37 participants in each condition (N = 111).
In order to compensate for possible drop-outs, we collected data of 122 participants using the Clickworker online platform\footnote{https://www.clickworker.com/clickworker/}.

To ensure a sufficient English level, participation was limited to users from the US, UK, Australia, or Canada whose native language is English.
Since LRP and LIME are not designed with color blind people in mind, the participants were also asked if they were color blind and stopped from participating if they are.

To make sure that the participants understood the provided information about the task correctly, we used a quiz that they had to complete correctly to take part in the study.
As incentive to diligently do the task, the participants received a bonus payment in addition to the base payment if they correctly predicted at least $2/3$ of the AI model's prediction.
In addition to these precautions, we subsequently excluded 4 participants due to the fact that they never looked at the XAI visualisations or their responses did not reflect a serious engagement with the study (e.g., free text answers which are not related to the question at all).

For our final analysis we used data from 118 participants between 18 and 67 years (\textit{M}~=~38.5, \textit{SD}~=~10.9). 63 of them were male, 53 female and 2 non-binary.
The participants were randomly separated in the three XAI visualisation conditions. 
All in all, only 8 participants reported experience in health care. 43 participants stated that they had experience in AI. The level of AI and healthcare experience was evenly distributed between the three conditions.

\subsection{Procedure}
The entire study was web-based.
After providing some demographic information, the participants received a short tutorial that explained the x-ray images and the XAI visualisations which they would interact with in the experiment.
After the tutorial, each participant had to answer a quiz.
Here, questions were asked to ensure that the participants carefully read the tutorial and understood how to interpret the x-ray images (e.g., ``Which part of the body is marked in this picture?'') and the XAI visualisations (e.g., ``What do green areas in images tell you?'' for the LIME and LRP conditions).
Only participants who solved the quiz successfully were allowed to participate in the actual experiment.

After the quiz followed the prediction task. 
Here, the participants were asked to predict the AI's diagnosis for $12$ different images.
To avoid cherry picking while still ensuring variety in the images, we randomly chose $12$ images based on the following constraints: 
To make sure that the classifier equally makes false and correct predictions for each class, we wanted $3$ true positives, $3$ false positives, $3$ true negatives, and $3$ false negatives.
Furthermore, inspired by \cite{alqaraawi2020}, we additionally used the AI model's confidence to ensure diversity in the images. 
Decisions where the model is certain are often easier to interpret than decisions where the AI model struggled.
Since our prediction classifier mainly had confidence values between $0.8$ and $1$, we randomly chose one x-ray image with confidence values of $0.8$, $0.9$ and $1$ (rounded) out of each of the sets of true positives, false positives, true negatives, and false negatives.

\begin{figure}[t]
    \centering
    \begin{minipage}{0.24\linewidth}
    \centering
        \includegraphics[width=\linewidth]{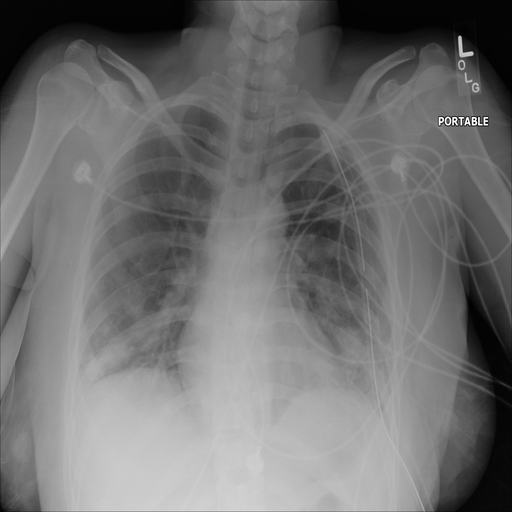}
        \parbox[t][0.5\baselineskip]{\linewidth}{
        \centering
        Original Input}
    \end{minipage}
    \begin{minipage}{0.24\linewidth}
    \centering
        \includegraphics[width=\linewidth]{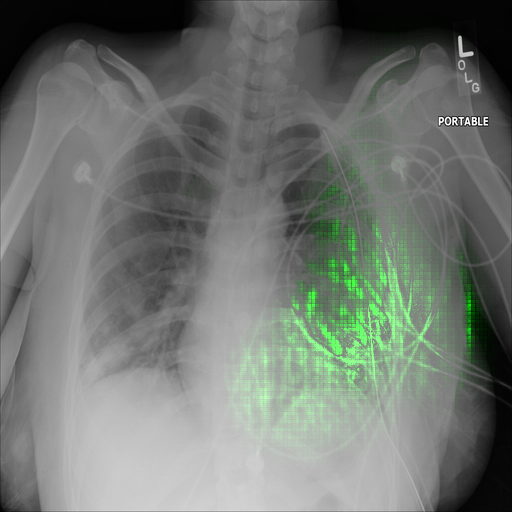}
        \parbox[t][0.5\baselineskip]{\linewidth}{\centering LRP}
    \end{minipage}
    \begin{minipage}{0.24\linewidth}
    \centering
        \includegraphics[width=\linewidth]{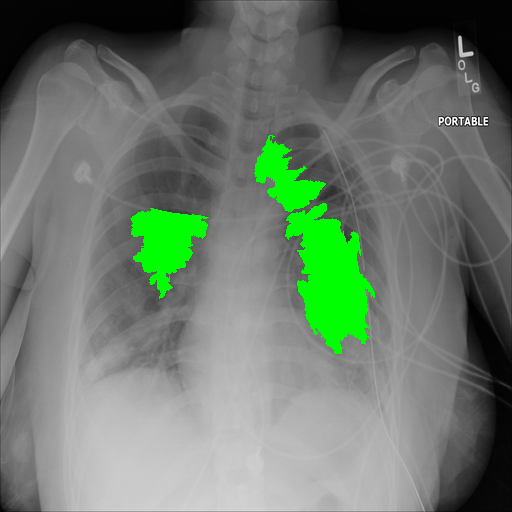}
        \parbox[t][0.5\baselineskip]{\linewidth}{\centering LIME}
    \end{minipage}
    \begin{minipage}{0.24\linewidth}
    \centering
        \includegraphics[width=\linewidth]{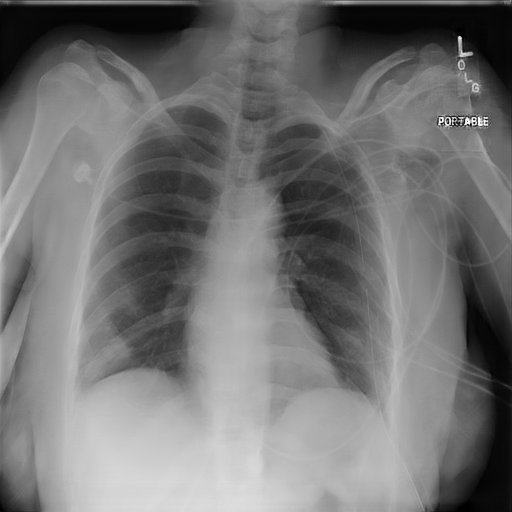}
        \parbox[t][0.5\baselineskip]{\linewidth}{\centering Counterfactual}
    \end{minipage}
    \caption{An example x-ray image classified as \textit{Pneumonia}, as well as the different XAI  visualizations used in our study when the slider is fully on the right side. Best viewed in color.}
    \label{fig:XAI_vis_study}
\end{figure}

In addition to the original image, the participants were provided with a slider to interact with the XAI visualizations.
Moving the slider to the right linearly interpolated the original image with either the counterfactual image or a version of the image that is augmented with a LRP or LIME heatmap, depending on the condition of the user.
Figure \ref{fig:XAI_vis_study} shows an example of the $3$ different XAI visualizations for one of the images used in our experiment.
By tracking if the participants used the slider, we additionally know whether they looked at the XAI visualizations.

\begin{figure}
    \centering
    \includegraphics[width=1\linewidth]{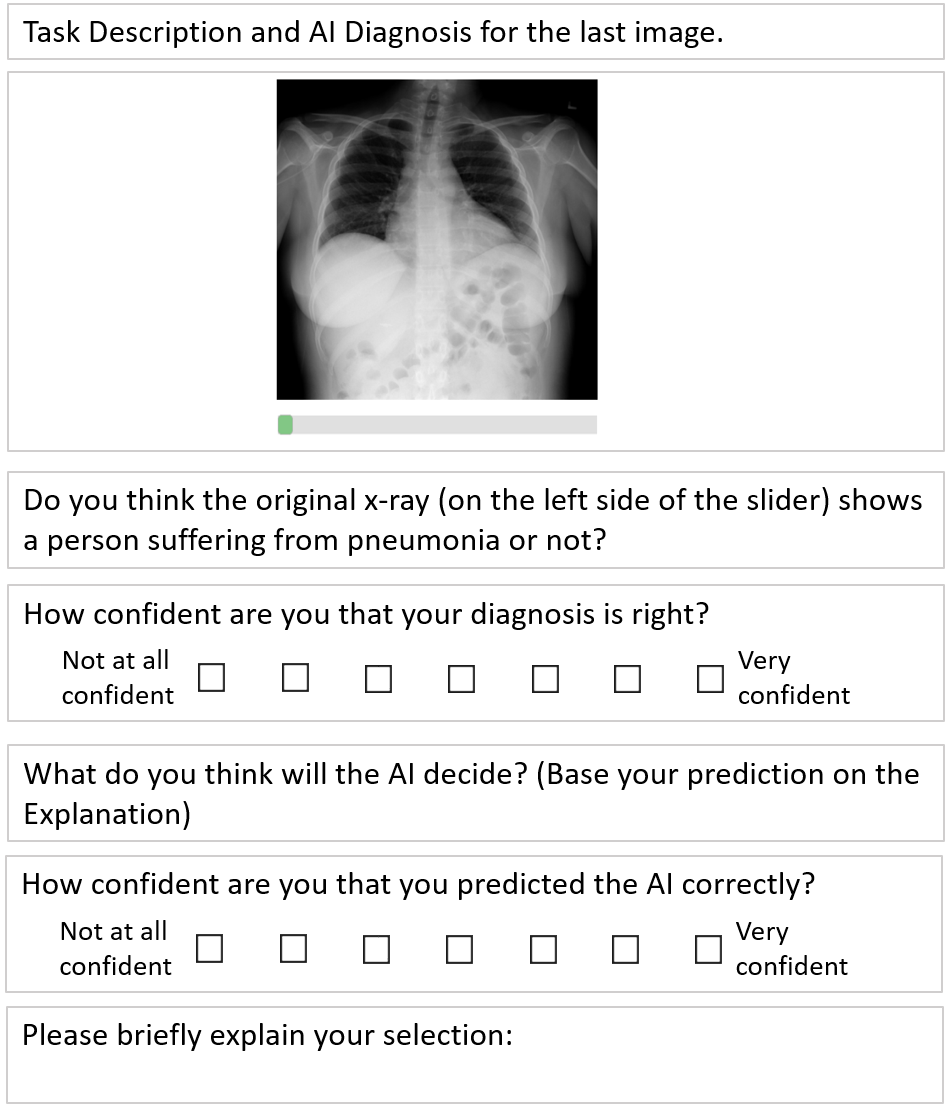}
    \caption{A simplified schematic of our prediction task.}
    \label{fig:prediction_task}
\end{figure}

We found in our pilot study (N = 10) that participants often project their own reasoning on the AI.
To mentally differentiate between their own diagnosis and the AI's diagnosis, the participants in the final study were asked whether they \emph{themselves} would classify the given image as \emph{pneumonia} or \emph{not pneumonia} and how confident they are in this diagnosis on a Likert scale from $1$ (not at all confident) to $7$ (very confident).
Then they were asked to predict whether \emph{the AI} will classify the image as \emph{pneumonia} or \emph{not pneumonia}, based on the given XAI visualization. 
Here too, they had to give a confidence rating in their prediction from $1$ to $7$.
Finally, they could give a justification for their prediction if they wanted to.
After each prediction they were told the actual decision of the AI for the last image.
A schematic of the full task is shown in Figure \ref{fig:prediction_task}.

After predicting the AI's decision for all $12$ x-ray images followed the task reflection where they had to describe their understanding of the AI's reasoning.
Then the questionnaires about Explanation Satisfaction, Trust, Self-efficacy and Emotion were provided.

\subsection{Evaluation Methods}

\paragraph{Quantitative Evaluation of the Results}
We calculated the mean of the correct predictions of the AI and the participants confidences in their predictions of the AI. 
To make sure that we only use responses, where the participants at least saw the visual explanations, we excluded answers where the participant did not move the slider.
If, for example, a participant did not use the slider $4$ times then we only calculated the mean for the remaining $8$ answers.

For the DEQ we calculated the mean for the emotion subscales happy, anger, and relaxation. For the TiA, we calculated an overall trust score from the two questions presented.

\paragraph{Qualitative Evaluation of the Participants' Mental Model of the AI}
Similar to \cite{anderson2019mere-mortals} and \cite{huber2020LocalandGlobal}, we used a form of summative content analysis \cite{hsieh2005content_analysis} to qualitatively evaluate the participants' free text answers to the questions ``What do you think the AI pays attention to when it predicts pneumonia?'' and ``What do you think the AI pays attention to when it predicts healthy lungs?''.
Our classifier was trained on a dataset consisting of x-ray images of normal lungs and x-ray images that contain lung opacity, which is a crucial indicator of lungs suffering from pneumonia.
Since we only told the participants that our model classifies pneumonia, we can score their responses based on whether they correctly identified lung opacity as a key decision factor for our model. 
To this end, two annotators independently went through the answers and assigned concepts to each answer (e.g. \textit{opacity}, \textit{clarity}, \emph{contrast} and \emph{other organs than the lung}).
Then, answers to the pneumonia question that contained at least one concept which related to opacity, like \emph{opacity}, \emph{white color in the x-ray} and \emph{lung shadows}, received $1$ point. 
Answers to the healthy lungs question that contained at least one concept related to clarity, like \emph{clarity}, \emph{black color in the x-ray} or \emph{no lung shadows}, received $1$ point.
Answers for both questions that contained a concept related to contrast, like \emph{contrast} or \emph{clear edges}, received $0.5$ points.
All other answers received $0$ points.
For $21$ out of all $236$ responses, the two annotators differed in the given score.
Here, a third annotator was asked to assign $0$, $0.5$ or $1$ points to the answer and the final points were calculated by majority vote between the three annotators.
By adding the points for those two questions, each participant was given a score between $0$ and $2$ approximating the correctness of their description of the AI.

\section{Results}\label{sec:results}

\subsection{Impact of XAI methods on Explanation Satisfaction, Trust, and Prediction Accuracy}

\begin{figure}[t]
    \centering
    \begin{minipage}{0.49\linewidth}
    \centering
        \includegraphics[width=\linewidth]{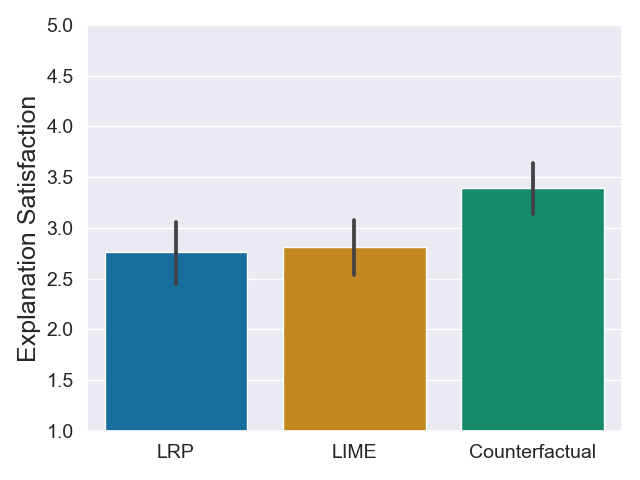}
        \parbox[t][0.5\baselineskip]{\linewidth}{
        \centering
        Expl. Satisfaction}
    \end{minipage}
    \begin{minipage}{0.49\linewidth}
    \centering
        \includegraphics[width=\linewidth]{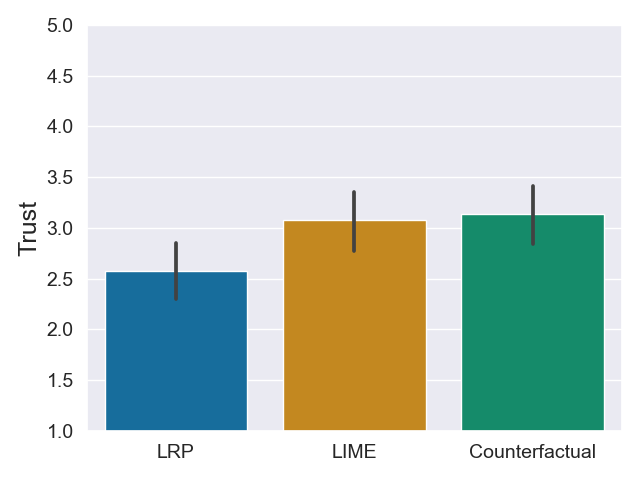}
        \parbox[t][0.5\baselineskip]{\linewidth}{\centering Trust}
    \end{minipage}
    \caption{Results of the explanation satisfaction and trust questionnaires. Error bars represent the 95\% Confidence Interval (CI).}
    \label{fig:results_main}
\end{figure}

\begin{figure}[t]
    \centering
    \begin{minipage}{0.49\linewidth}
    \centering
        \includegraphics[width=\linewidth]{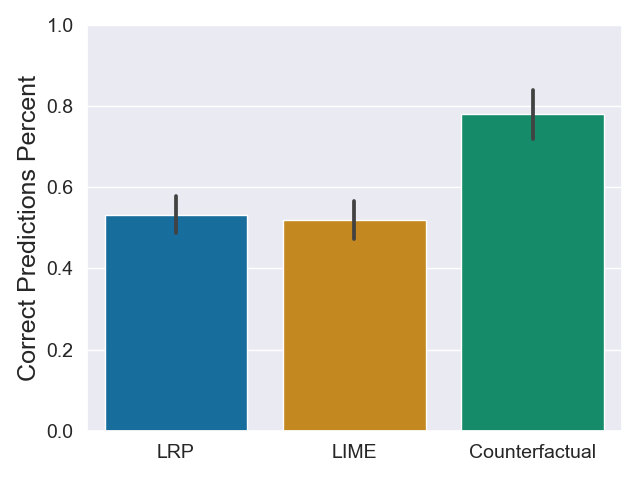}
        \parbox[t][0.5\baselineskip]{\linewidth}{\centering Prediction Accuracy}
    \end{minipage}
    \begin{minipage}{0.49\linewidth}
    \centering
        \includegraphics[width=\linewidth]{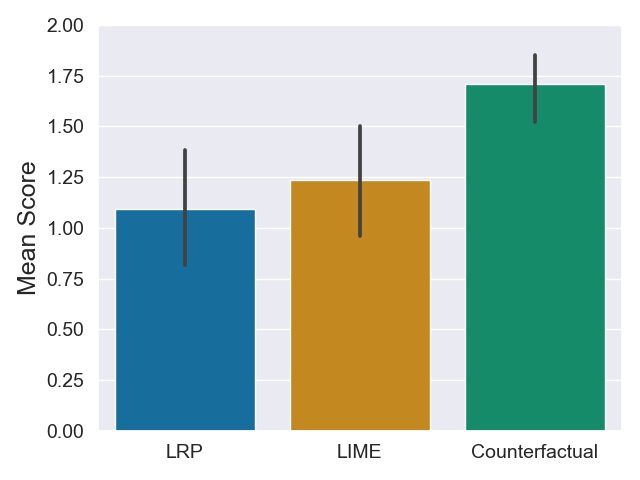}
        \parbox[t][0.5\baselineskip]{\linewidth}{\centering Task Reflection Score}
    \end{minipage}
    \caption{Results of the prediction task, and the task reflection questions. Error bars represent the 95\% Confidence Interval (CI).}
    \label{fig:results_mental_model}
\end{figure}

As a first impression of their mental models of the AI, the participants had to predict the decision of the neural network (pneumonia / no pneumonia).
At the end of the study, they rated their trust in the AI as well as their explanation satisfaction.
To evaluate these variables between the three conditions, we conducted a one-way MANOVA. Here we found a significant statistical difference, Wilks' Lambda~=~0.59, \textit{F}(6,~226)~=~11.2, \textit{p}~$<$~.001. The following ANOVA revealed that all three variables showed significant differences between the conditions:
\begin{itemize}
    \item \textbf{Prediction accuracy}: \textit{F}(2,~115)~=~30.18, \textit{p}~=~.001,
    \item \textbf{Explanation satisfaction}: \textit{F}(2,~115)~=~5.87, \textit{p}~=~.004,
    \item \textbf{Trust}: \textit{F}(2,~115)~=~3.89, \textit{p}~=~.02,
\end{itemize} 
To determine the direction of the differences between the three XAI method conditions, we used post-hoc comparisons for each variable\footnote{We used the Holm correction for multiple testing to adjust the p-values for all post-hoc tests we calculated.}. We found the following differences: 
\begin{itemize}
    \item \textbf{Prediction accuracy}:  
    The participants' predictions of the AI's decisions were significantly more correct in the counterfactual condition compared to the LRP condition \textit{t}~=~-6.48, \textit{p}~=~.001, \textit{d}~=~1.47 (large effect) as well as compared to the LIME conditions \textit{t}~=~-.92, \textit{p}~=~.001, \textit{d}~=~1.56 (see left sub-figure of Figure \ref{fig:results_mental_model}).
    \item \textbf{Explanation satisfaction}: Participants were significantly more satisfied with the explanation quality of the counterfactual explanations compared to the LRP saliency maps, \textit{t}~=~-3.05, \textit{p}~=~.008, \textit{d}~=~0.70 (medium effect) and the LIME visualisations, \textit{t}~=~-2.85, \textit{p}~=~0.01, \textit{d}~=~0.64 (medium effect)(see Figure \ref{fig:results_main}).
    \item \textbf{Trust}: The AI was rated as significantly more trustworthy in the counterfactual condition compared to the LRP condition, \textit{t}~=~-2.56, \textit{p}~=~.03, \textit{d}~=~0.58 (medium effect) but not to the LIME condition, \textit{t}~=~-0.29, \textit{p}~=~.07 (see Figure \ref{fig:results_main}).
\end{itemize}

\subsection{Result of the qualitative Evaluation of the Users' Mental Models}

Subsequently to the significant differences in the prediction accuracy as a first impression of the mental model of the participants, we analysed the results of the content analysis of the task reflection responses. 
For this, we conducted a one-way ANOVA.
Here we found a significant statistical difference, \textit{F}(2,~115)~=~7.91, \textit{p}~$<$~.001. 
To determine the direction of the differences between the three conditions, we used post-hoc comparisons (see right sub-figure of Figure \ref{fig:results_mental_model}):
Participants were asked to describe the AI's reasoning in three different conditions: counterfactual, LRP and LIME.
Out of these, participants created correct descriptions significantly more often in the counterfactual condition compared to the LRP condition, \textit{t}~=~-3.76, \textit{p}~$<$~.001, \textit{d}~=~0.85 (large effect) and the LIME condition, \textit{t}~=~-2.97, \textit{p}~=~0.01, \textit{d}~=~0.66 (medium effect).

\subsection{Impact of XAI Methods on Users' Emotional State}

\begin{figure}[t]
    \centering
    \begin{minipage}{0.32\linewidth}
    \centering
        \includegraphics[width=\linewidth]{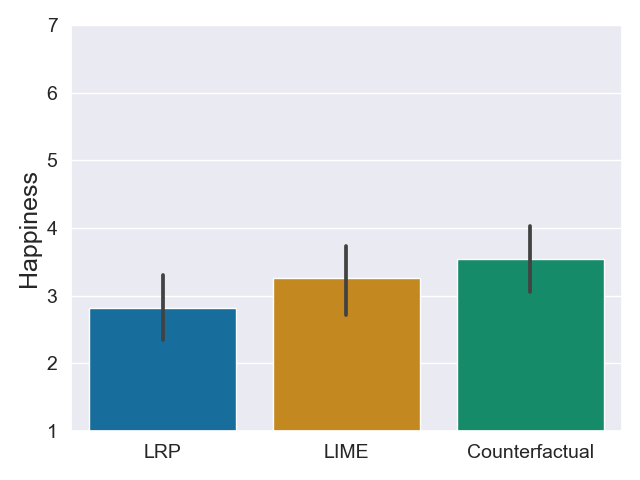}
        \parbox[t][0.5\baselineskip]{\linewidth}{
        \centering
        Happiness}
    \end{minipage}
    \begin{minipage}{0.32\linewidth}
    \centering
        \includegraphics[width=\linewidth]{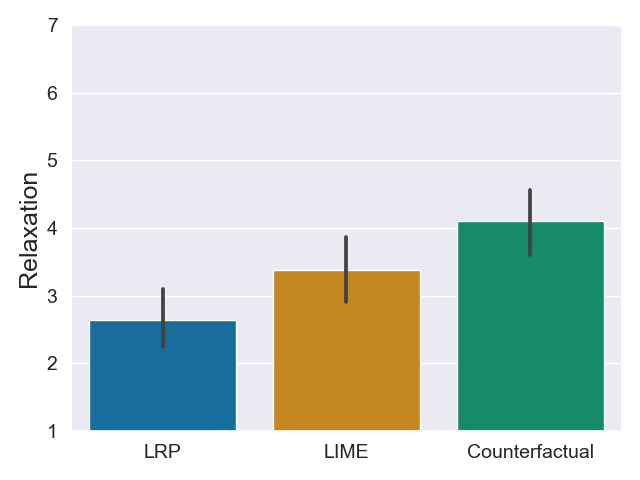}
        \parbox[t][0.5\baselineskip]{\linewidth}{\centering Relaxation}
    \end{minipage}
    \begin{minipage}{0.32\linewidth}
    \centering
        \includegraphics[width=\linewidth]{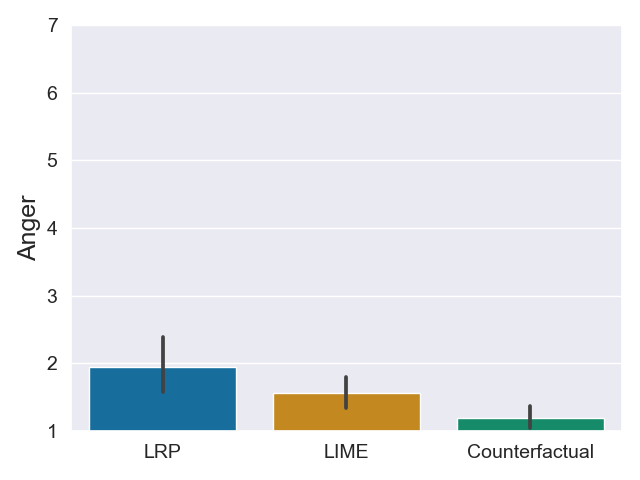}
        \parbox[t][0.5\baselineskip]{\linewidth}{\centering Anger}
    \end{minipage}
    \caption{Results of the emotion questionnaires. Participants in the counterfactual condition felt significantly less angry and more relaxed compared to the LRP saliency map condition. For LIME, no significant differences were found. Error bars represent the 95\% CI.}
    \label{fig:results_emotion}
\end{figure}

We also wanted to investigate whether working with the XAI methods had an influence on the emotional state of the participants.
To analyse possible effects, we conducted a one-way MANOVA. 
Here we found a significant statistical difference, Pillai's' Trace~=~0.20, \textit{F}(6,~228)~=~4.26, \textit{p}~$<$~.001. The following ANOVA revealed that the emotion anger, \textit{F}(2,~115)~=~5.68, \textit{p}~=~.002 and relaxation, \textit{F}(2,~115)~=~9.07, \textit{p}~$<$~.001 showed significant differences between the conditions. Happy showed no significant differences between the conditions, \textit{F}(2,~115)~=~2.06, \textit{p}~=~.13.
The post-hoc comparisons\footnote{We used the Holm correction for multiple testing to adjust the p-values} showed the following differences (see Figure \ref{fig:results_emotion}):
\begin{itemize}
    \item \textbf{Anger}: Participants in the counterfactual condition felt significantly less angry than in the LRP condition, \textit{t}~=~3.68, \textit{p}~=~.001, \textit{d}~=~0.83 (large effect). No differences were found for the LIME condition, \textit{t}~=~1.83, \textit{p}~=~.12, \textit{d}~=~0.86 (large effect)
    \item \textbf{Relaxation}: Participants in the counterfactual condition were significantly more relaxed than in the LRP condition, \textit{t}~=~-4.26, \textit{p}~$<$~.001., \textit{d}~=~0.96 (large effect). No differences were found for the LIME condition, \textit{t}~=~-2.12, \textit{p}~$<$~.06\footnote{This p-value was no longer significant due to the Holm correction.}
\end{itemize}

\subsection{Impact of XAI Methods on Users' Self-Efficacy}
\begin{figure}[t]
    \centering
    \begin{minipage}{0.49\linewidth}
        \centering
        \includegraphics[width=\linewidth]{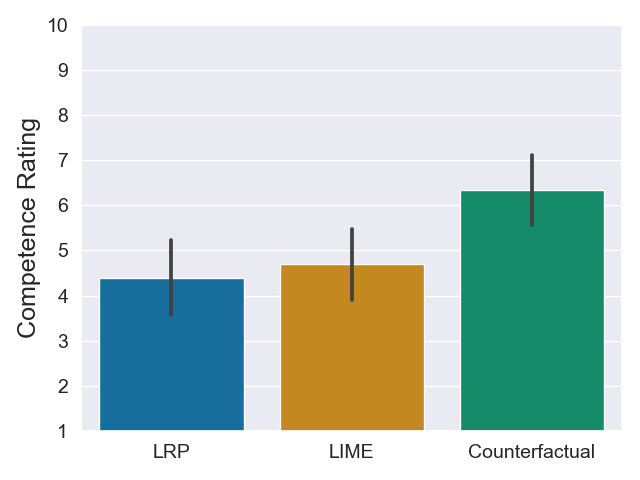}
        \parbox[t][0.5\baselineskip]{\linewidth}{\centering Self-Efficacy}
    \end{minipage}
    \begin{minipage}{0.49\linewidth}
        \centering
        \includegraphics[width=\linewidth]{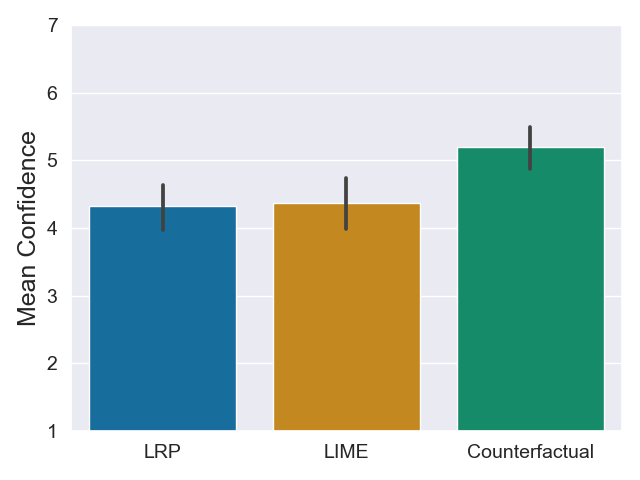}
        \parbox[t][0.5\baselineskip]{\linewidth}{\centering Confidence}
    \end{minipage}
    \caption{Significant differences regarding self-efficacy and general confidence of the participants in their predictions of the AI between the counterfactual condition and the saliency map conditions (LRP and LIME). Error bars represent the 95\% CI.}
    \label{fig:results_efficacy_generalconfidence}
\end{figure}

\begin{figure}[t]
    \centering
    \begin{minipage}{0.49\linewidth}
        \centering
        \includegraphics[width=\linewidth]{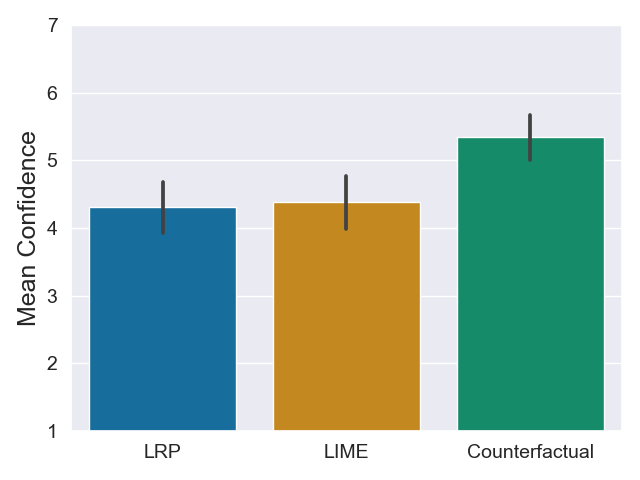}
        \parbox[t][0.5\baselineskip]{\linewidth}{\centering Correct Predictions}
    \end{minipage}
    \begin{minipage}{0.49\linewidth}
        \centering
        \includegraphics[width=\linewidth]{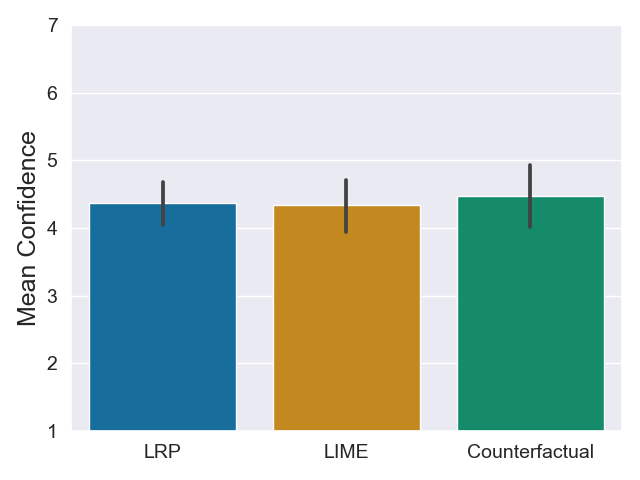}
        \parbox[t][0.5\baselineskip]{\linewidth}{\centering False Predictions}
    \end{minipage}
    \caption{Confidence of the participants in correct and false predictions. The significant difference between the counterfactual condition and the saliency map conditions is based on the confidence in correct predictions, not in the incorrect ones. Error bars represent the 95\% CI.}
    \label{fig:results_confidence_divided}
\end{figure}

The analysis showed that (1) the quality of counterfactual explanations was rated significantly higher and (2) participants predicted the decisions of the AI significantly more accurate compared to LIME and LRP.
%In addition to this we
Based on our last hypothesis, we therefore examined whether these positive assessments were also reflected in the self-efficacy and in the prediction confidence of the participants.
For this purpose, we conducted a one-way MANOVA. Here, we found a significant statistical difference, Pillai's Trace~=~0.15, \textit{F}(4,~230)~=~4.69, \textit{p}~$=$~.001. The following ANOVA revealed a statistical difference for self-efficacy \textit{F}(2,~115)~=~6.93, \textit{p}~$=$~.001 and prediction confidence \textit{F}(2,~115)~=~3.68, \textit{p}~$<$~.001 between the conditions.
The post-hoc comparisons showed that counterfactuals lead to a significantly higher self-efficacy compared to LRP \textit{t}~=~-3.44, \textit{p}~$=$~.002, \textit{d}~=~0.78 (medium effect) as well as LIME, \textit{t}~=~-2.94, \textit{p}~$=$~.01, \textit{d}~=~0.66 (medium effect). 
The same pattern was found for the prediction confidence, where counterfactuals lead to a significantly higher prediction confidence compared to LRP \textit{t}~=~-3.45, \textit{p}~$=$~.002, \textit{d}~=~0.78 (medium effect) as well as LIME, \textit{t}~=~-3.32, \textit{p}~$=$~.003, \textit{d}~=~0.74 (medium effect) (see Figure \ref{fig:results_efficacy_generalconfidence}). A closer look reveals that these significant differences stem from the confidence in the correct predictions and not the confidence in the incorrect ones (see Figure \ref{fig:results_confidence_divided}).

\section{Discussion}\label{sec:discussion}
The study described in the previous sections was conducted with the aim to verify our hypotheses. With this in mind, we discuss our results in this section.

\subsection{Explanation Satisfaction}
As the results show, the counterfactual explanation images that were generated by the use of our novel approach provided the participants with significantly more satisfying explanations as both of the saliency map approaches. 
Saliency map methods like LIME and LRP only show which pixels were important for the AI's decision.
The users are left alone with the task of building a bridge between the information of \textit{where} the AI looked at, and \textit{why} it looked there.
Contrary, the counterfactual explanations generated by our system directly show, \textit{how} the input image would have to be modified to alter the AI's decision.
Thus, the participants did not have to come up with an interpretation of the semantics of important areas by themselves. 
As the results of our study show, this difference plays a significant role in how satisfying the explanations are to non-expert users, validating our first hypothesis.

\subsection{Mental Models}
As described in section \ref{sec:study}, two different methods were used to evaluate if the explanation systems allowed the participants to build up an appropriate mental model of the classifier. 
First, the participants had to do a prediction task of 12 images, where they had to decide if the AI would classify each of those images either as \textit{Pneumonia} or \textit{No Pneumonia}.
Our results show that the participants were significantly better in performing those prediction tasks when they were shown counterfactual %explanation 
images created by our system than they were when provided with LIME or LRP saliency maps. 
Again, it could be argued that this advantage is caused by the fact that the counterfactual images give more than just a spatial information about the regions of importance.
In fact, the actual decision of the AI was highly dependent on the blurriness of certain areas of the lung.
A crucial thing to mention is that the absence of blurriness, i.e. the clarity of x-ray images that do not show lungs that are infected by pneumonia, obviously occurs at similar places where cloudy areas would appear in the case of pneumonia. 
Thus, the visual highlighting created by LIME or LRP predominantly shows where this distinction between opaque and not opaque lungs is made.
However, the information is missing to which degree the AI actually thinks that there is an opacity in the lung.
In contrast, the counterfactual images give this information by increasing or decreasing that opacity respectively.
In general, we think that our counterfactual system has the most advantage in these kind of tasks, where the important regions are not distinct for different decisions.
Specifically, we think that our approach excels in tasks where the AI's decision is being directed by different textural characteristics rather than by the position of certain objects in the image.
The content analysis of the task reflection strengthens this assumption.
Here, participants from the LRP and LIME conditions often referred to certain organs or regions in the image instead of focusing on the key decision factor of opacity.
Examples for this are: ``The AI pays attention not to just the lungs but the surrounding areas as well. The Abdomen seems to be an area of focus.'', ``From the heatmap I noticed the AI paying attention to the surrounding areas of the lungs, the spine, heart, abdomen, and the armpits often when it predicted pneumonia.'' and ``I think the AI needs to see the green near the bottom of the chest to think healthy lungs.''

\subsection{Trust}
Our results show that counterfactual explanations encouraged the participants to have more trust in the AI system.
However, this only became apparent in comparison to LRP, but not to LIME. 
This result indicates on the one hand that the type of explanation (counterfactual explanation vs. feature importance/saliency maps) has an influence on the perceived trust of users.
On the other hand, it also shows that even explanations of one XAI type (here: saliency map approaches) are perceived differently by users. This finding is important because it indicates that the type of visualisation (pixel-wise or superpixel-based) also has an influence on the users' trust rating. 
In our study we examined the general influences of three XAI methods on trust.
Based on the results, further analyses are now necessary.
For example, the question arises whether there is a correlation between the participants' predictions and the trust rating.
One interesting observation in our results is that participants in the LIME condition trusted the system on a similar level as the participants in counterfactual condition even though they did significantly worse in the mental model evaluation.
This indicates that their trust might not be justified.
While this is interesting, the question of whether the trust of the participants in the AI system was actually justified needs to be examined more closely in the future. 

\subsection{Emotions}
In our user study, we not only investigated the impact of XAI visualisations on trust and mental models, but also for the first time the emotional state of the participants. The result shows that XAI not only influences users' understanding and trust, but also has an impact on users' affective states. Counterfactual explanations promote positive emotions (i.e., relaxation) and reduce negative emotions (i.e., anger). Kaptein et al. \cite{kaptein2017role} argue in their paper that emotions should be included as an important component of AI self-explanations (e.g., self-explanatory cognitive agents). Based on our results, we extend this argument by stating that users' emotions should also be taken into account in XAI designs.

\subsection{Self-efficacy}
Our results show that participants were not only able to correctly assess the predictions of the AI with the help of the counterfactual explanations, but also that they were very confident in their judgements. 
Upon closer inspection we found that this boost in confidence only stems from the predictions which the participants got right.
This indicates that they were not overconfident but justified in their confidence. 
While this is an interesting observation, it needs further investigation.
The increase in confidence is also reflected in a significant increase in the self-efficacy of participants in the counterfactual condition, compared to LIME and LRP.
Already \cite{heimerl2020unraveling} assumed that the use of XAI could be a valuable support to improve self-efficacy towards AI.
This assumption was empirically proven for the first time in our study and contributes to towards a more human-centred AI.

\subsection{Limitations}
It has to be investigated further how our proposed counterfactual generation method performs in other use cases. 
We believe that the advantage of our system in this pneumonia detection scenario to some degree results from the fact that the relevant information of the images is of a rather textural structure.\\
A further noteworthy observation is that, although the study showed that the produced counterfactuals lead to good results in our chosen non-expert task, our system modifies relevant features in a very strong way, i.e. features that are relevant for the classifier are modified to such a degree that the classifier is \textit{sure} that the produced image belongs to the respective other class. 
As these strong image modifications point out the relevant features in a very emphasized way, they lead to satisfactory explanations for non-experts that are not familiar with fine details of the problem domain. 
However, those kind of explanations might not be optimal for expert users, as those could perceive the performed feature translation as an exaggerated modification of the original features. The adaption of our system for an expert system would demand for further modification of our proposed loss function to produce images that are closer to the classifier's decision boundary.
We already propose a possible adjustment for this in section \ref{sec:approach} but did not test this adjustment thoroughly yet.\\ 
In our work, we presented a use case that was based on a binary classification problem. We want to emphasize that the proposed method can in theory easily be extended to a multi-class classification problem. In order to do so, multiple CycleGAN models have to be trained. When dealing with $k$ classes $\{S_1, ..., S_k\}$, for every pair of classes $(S_i,S_j)$, with $i \neq j$,
a CycleGAN has to be trained to solve the translation task between domain $S_i$ and $S_j$, resulting in $\frac{k!}{2(k - 2)!}$ models. Thus, the number of models is $O(k^2)$. While there is conceptually not a problem with this, the training of a huge number of models in practice can become a challenge due to limited resources. Thus, we see the application of our approach rather in explaining classifiers that do not deal with too many different classes.

\section{Conclusion and Outlook}\label{sec:conclusion}
In this paper, we introduced a novel approach for generating counterfactual explanations for explaining image classifiers.\\
Our computational comparison between counterfactuals generated by an original CycleGAN and a CycleGAN that was modified by our approach showed that our introduced loss component forces the model to predominantly generate images that were classified in a different way than the original input, while the original CycleGAN performed very poorly in this respective task. Thus, the introduced modification had a substantially positive impact generating counterfactual images.\\
Furthermore, we conducted a user study to evaluate our approach and compare it to two state-of-the-art XAI approaches, namely LIME and LRP.
As evaluation use case, we chose the explanation of a classifier that distinguishes between x-ray images of lungs that are infected by pulmonia and lungs that are not infected.
In this particular use case, the counterfactual approach outperformed the common XAI techniques in various regards.
Firstly, the counterfactual explanations that were generated by our system led to significantly more satisfying results as the two other systems that are based on saliency maps. 
Secondly, the participants formed significantly better mental models of the AI based on our counterfactual approach than on the two saliency map approaches.
Also, participants had more trust in the AI after being confronted with the counterfactual explanations than with the LRP condition.
Furthermore, users that were shown counterfactual images felt less angry and more relaxed than users that were shown LRP images.\\
All in all, we showed that our approach is very promising and shows great potential for being applied in similar domains.\\
However, it has to be investigated further how the system performs in other use cases. 
We believe that the advantage of our system in this specific scenario results from the relevant information of the images being of a rather textural structure, e.g. opacity. 
Thus, raw spatial information about important areas, as provided by LIME and LRP, do not carry enough information to understand the AI's decisions. Therefore, we recommend the application of our approach in similar use cases, where relevant class-defining features are expected to have a textural structure. 
To validate this hypothesis, we plan to conduct further research to evaluate our approach in different use cases. 

\section*{Acknowledgements}
This work has received funding from the DFG under project number 392401413, DEEP.\\
Further this work presents and discusses results in the context of the
research project ForDigitHealth. The project is part of the Bavarian Research
Association on Healthy Use of Digital Technologies and Media
(ForDigitHealth), funded by the Bavarian Ministry of Science and Arts.
Finally, we would like to thank our students Dominik Horn, Simon Kostin, Tobias Schmidt, Henrik Wachowitz, and Alexander Zellner, who assisted us with large parts of our systems implementation.

\bibliography{main}
\clearpage
\appendix
\section{Classifier Architecture}\label{appendix:classifier_architecture}

\begin{table}[h!]
\centering
\begin{tabular}{||c | l| c| c| c| c||} 
 \hline
Layer & Description & Number of Filters & Size & Stride & Dropout Probability\\ [0.5ex] 
 \hline\hline

1 & Conv2D & 96 & 11 x 11 & 4 & - \\\hline
2 & MaxPooling2D & - & 2 x 2 & 2 & -\\\hline
3 & Batch Normalization & - & - & - & -\\\hline
4 & Conv2D & 256 & 11 x 11 & 1 & -\\\hline
5 & MaxPooling2D & - & 2 x 2 & 2 & -\\\hline
6 & Batch Normalization & - & - & - & -\\\hline
7 & Conv2D & 384 & 3 x 3 & 1 & -\\\hline
8 & Batch Normalization & - & - & - & -\\\hline
9 & Conv2D & 384 & 3 x 3 & 1 & -\\\hline
10 & Batch Normalization & - & - & - & -\\\hline
11 & Conv2D & 256 & 3 x 3 & 1 & -\\\hline
12 & MaxPooling2D & - & 2 x 2 & 2 & -\\\hline
13 & Batch Normalization & - & - & - & -\\\hline
14 & Flatten & - & - & - & -\\\hline
15 & Dense & - & 4096 & - & -\\\hline
16 & Dropout & - & - & - & 0.4 \\\hline
17 & Batch Normalization & - & - & - & -\\\hline
18 & Dense & - & 4096 & - & -\\\hline
19 & Dropout & - & - & - & 0.4 \\\hline
20 & Batch Normalization & - & - & - & -\\\hline
21 & Dense & - & 1000 & - & -\\\hline
22 & Dropout & - & - & - & 0.4 \\\hline
23 & Batch Normalization & - & - & - & -\\\hline
24 & Dense & - & 2 & - & -\\ [1ex] 
 \hline
\end{tabular}
\label{table:classifier_architecture}
\caption{
L2 bias and kernel regularization with a regularization factor of 0.001 was applied to all convolutional and dense layers except layer 25.}
\end{table}

\section{Study Design}\label{appendix:study_design}
The following figures show the online study that was conducted. The condition of our counterfactual approach is depicted. The other conditions, i.e. LIME and LRP, were designed analogously. 

\begin{figure}[H]
 \centering
  \includegraphics[width=1\linewidth]{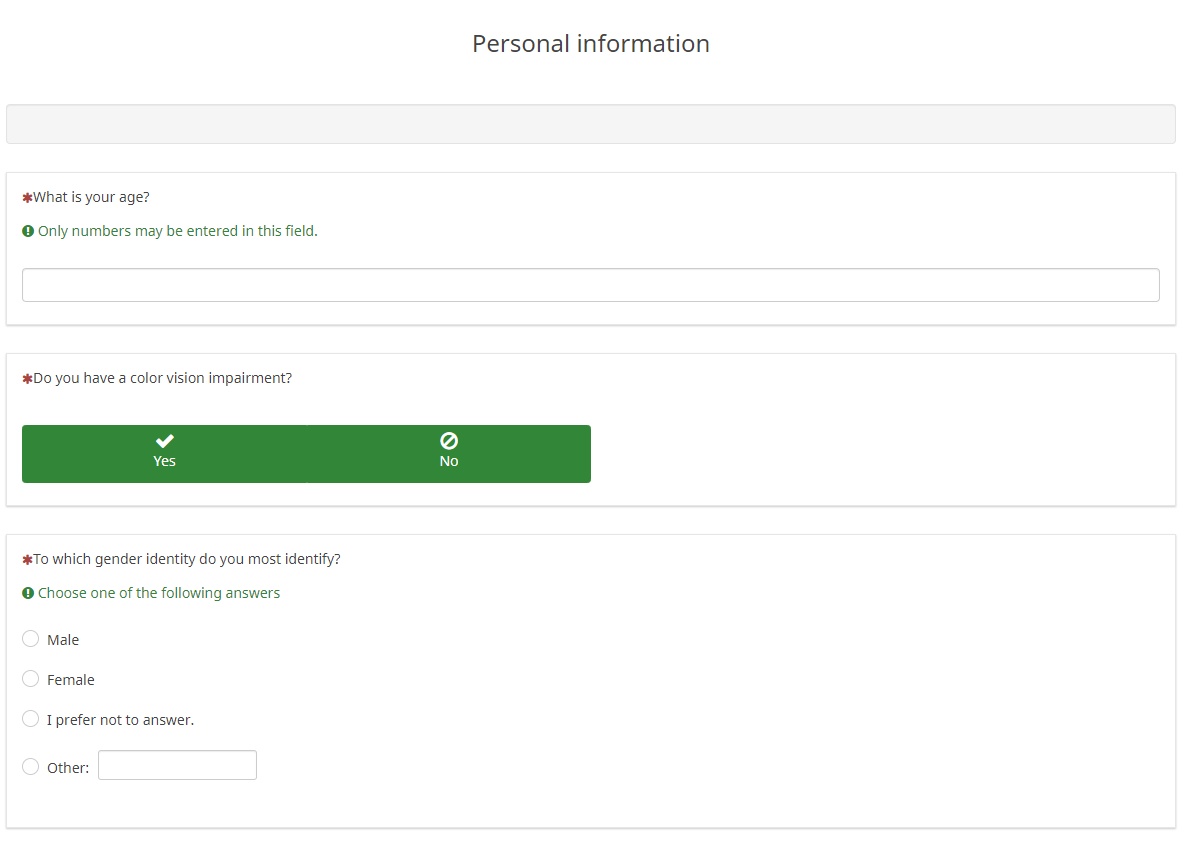}
  \caption{Study Design}
\end{figure}
\begin{figure}[H]
 \centering
  \includegraphics[width=1\linewidth]{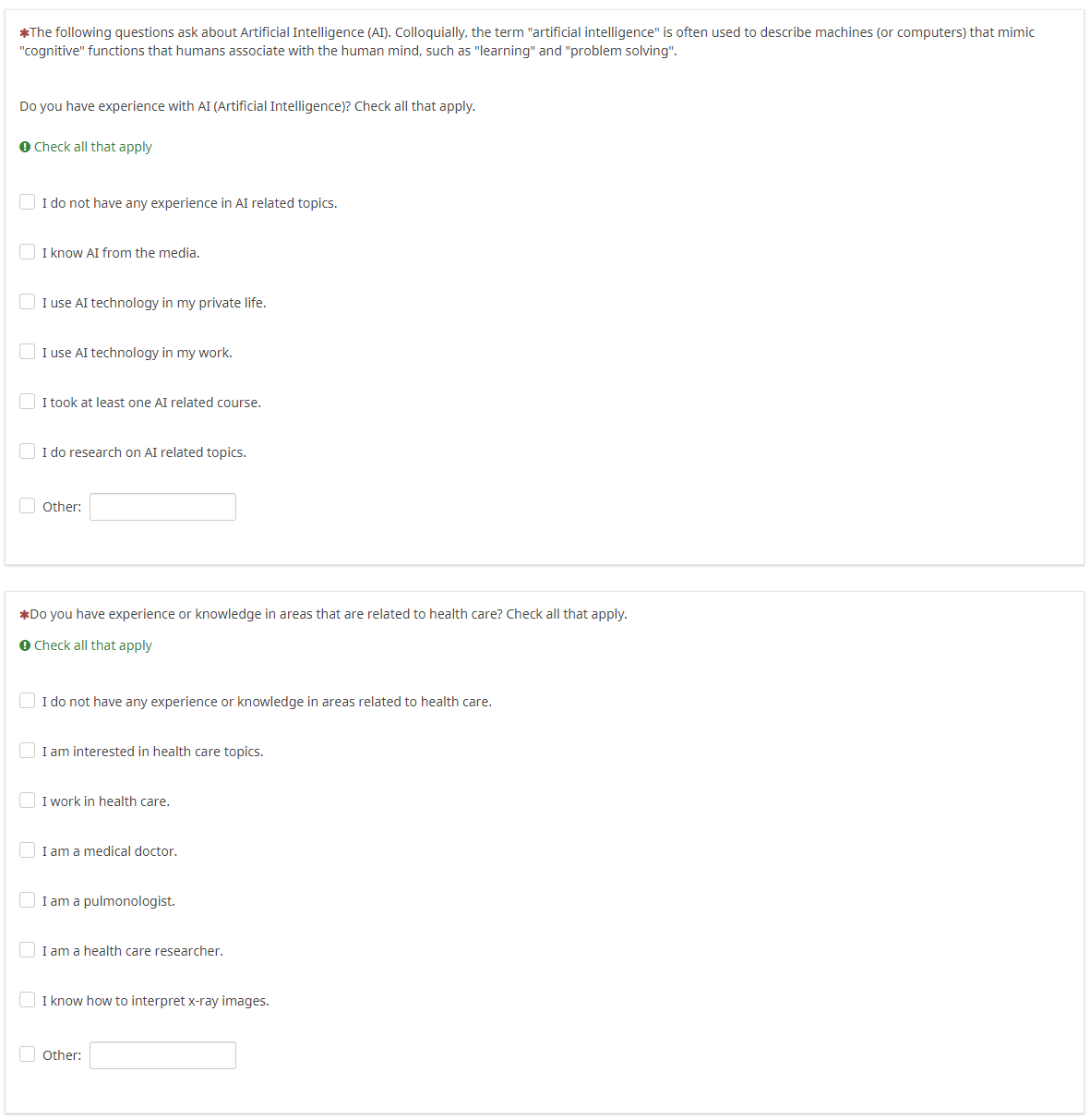}
  \includegraphics[width=1\linewidth]{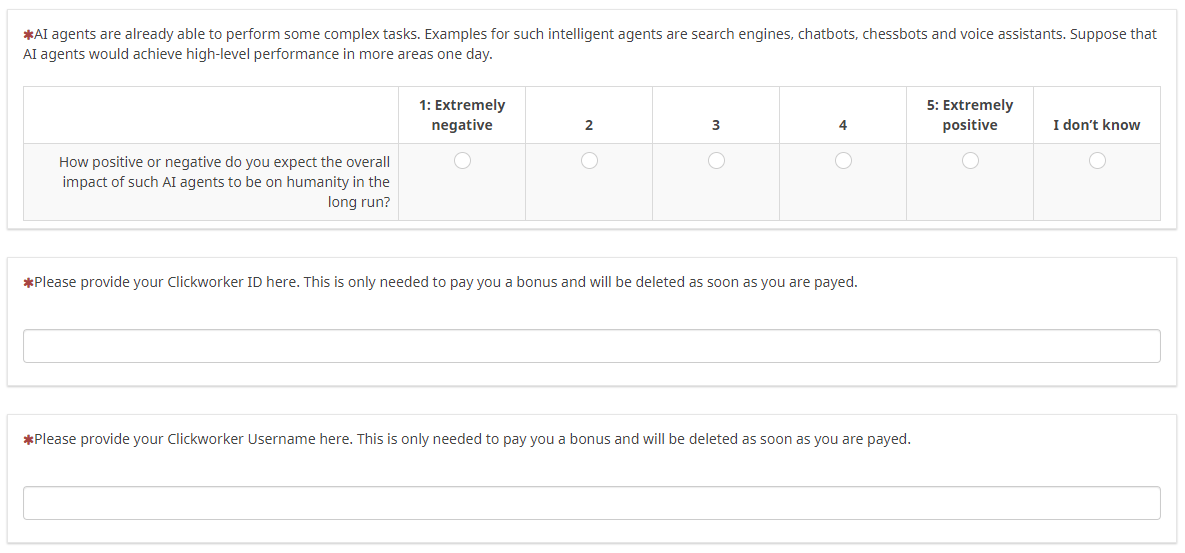}
\caption{Study Design}
\end{figure}

\begin{figure}[H]
 \centering
  \includegraphics[width=1\linewidth]{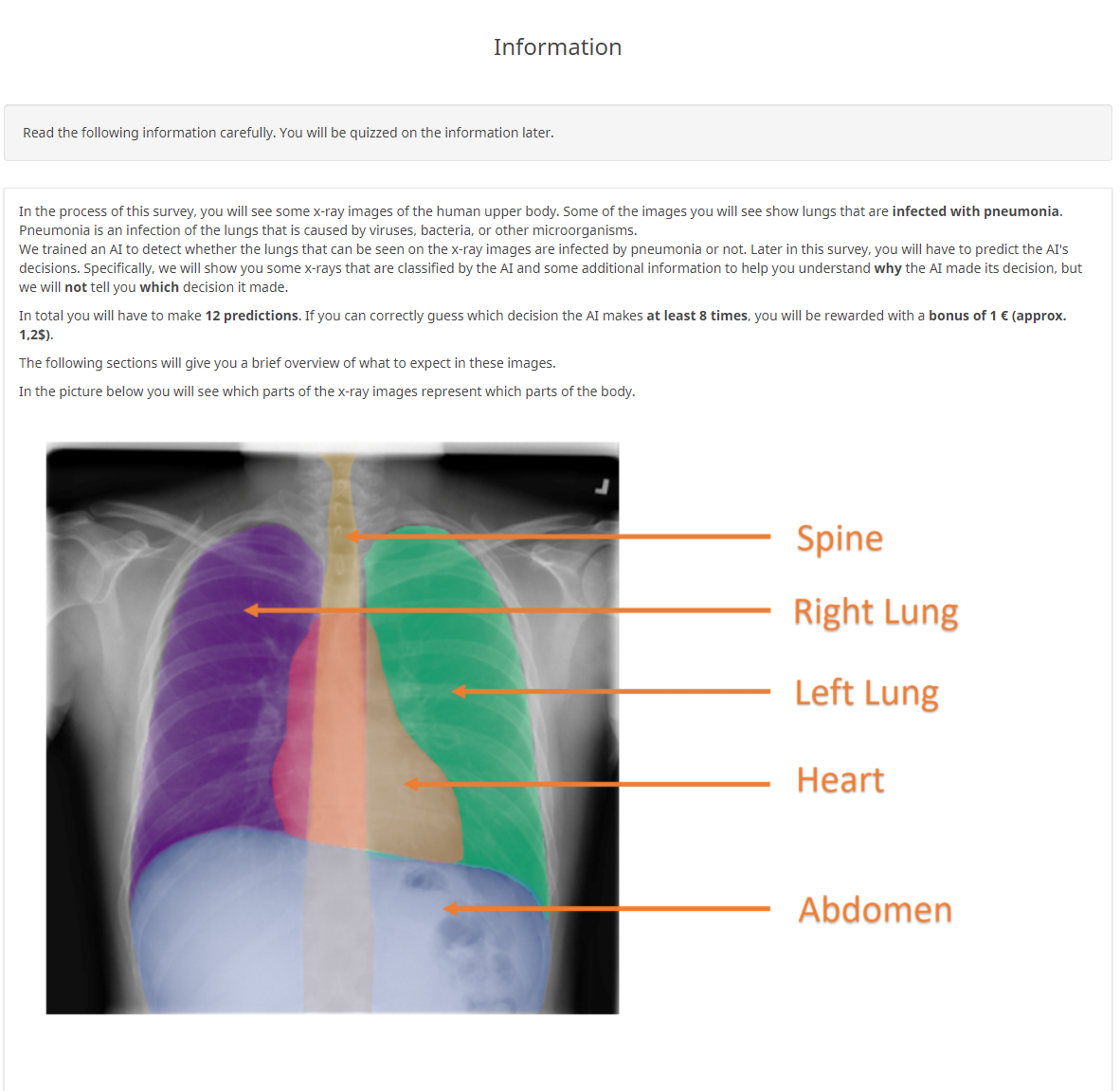}
  \includegraphics[width=1\linewidth]{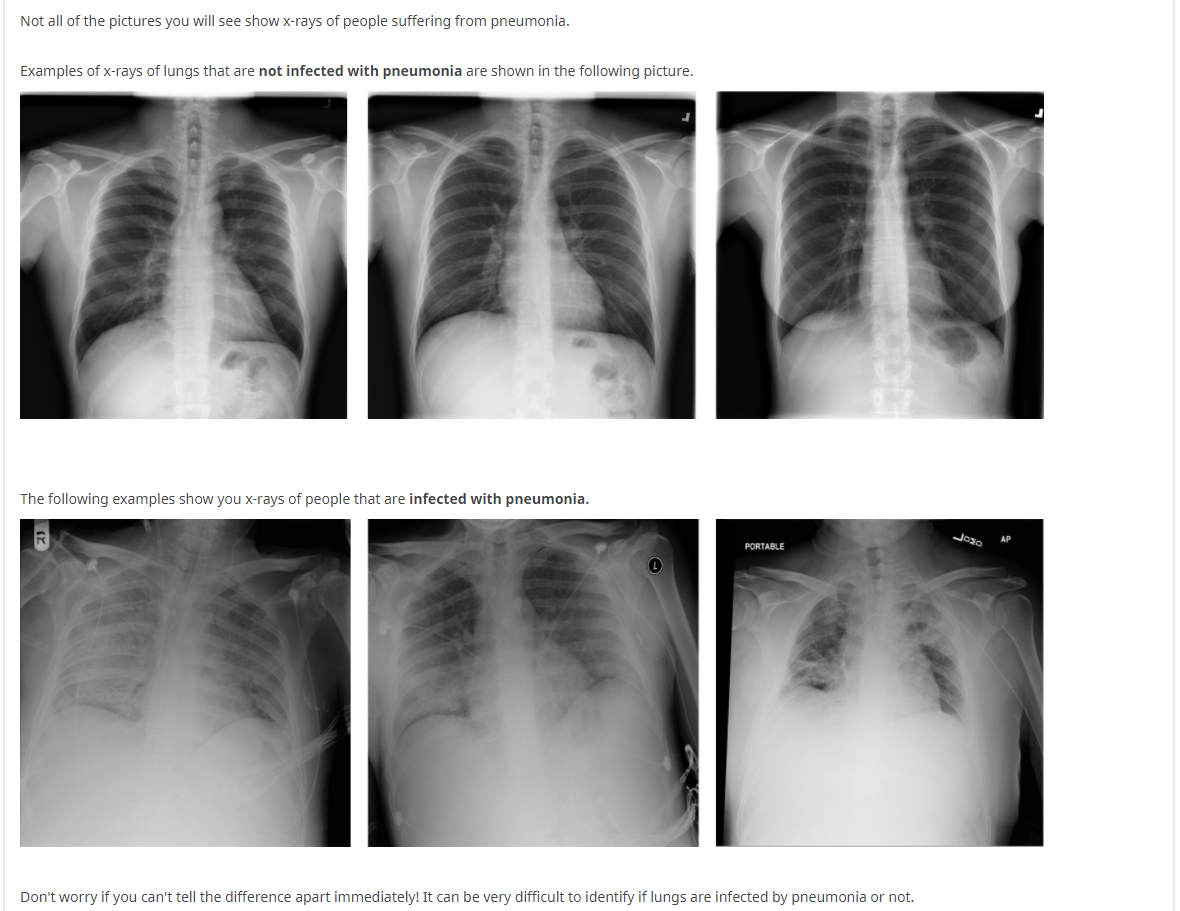}
\caption{Study Design}
\end{figure}

\begin{figure}[H]
 \centering
  \includegraphics[width=1\linewidth]{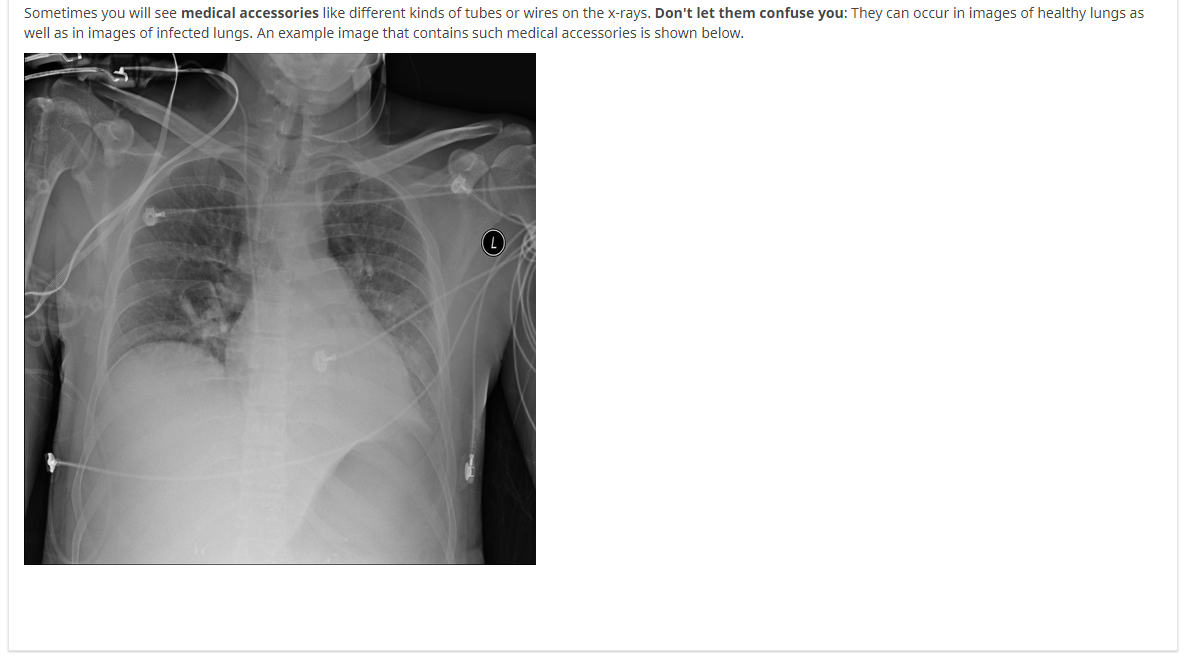}
\caption{Study Design}
\end{figure}

\begin{figure}[H]
 \centering
  \includegraphics[width=1\linewidth]{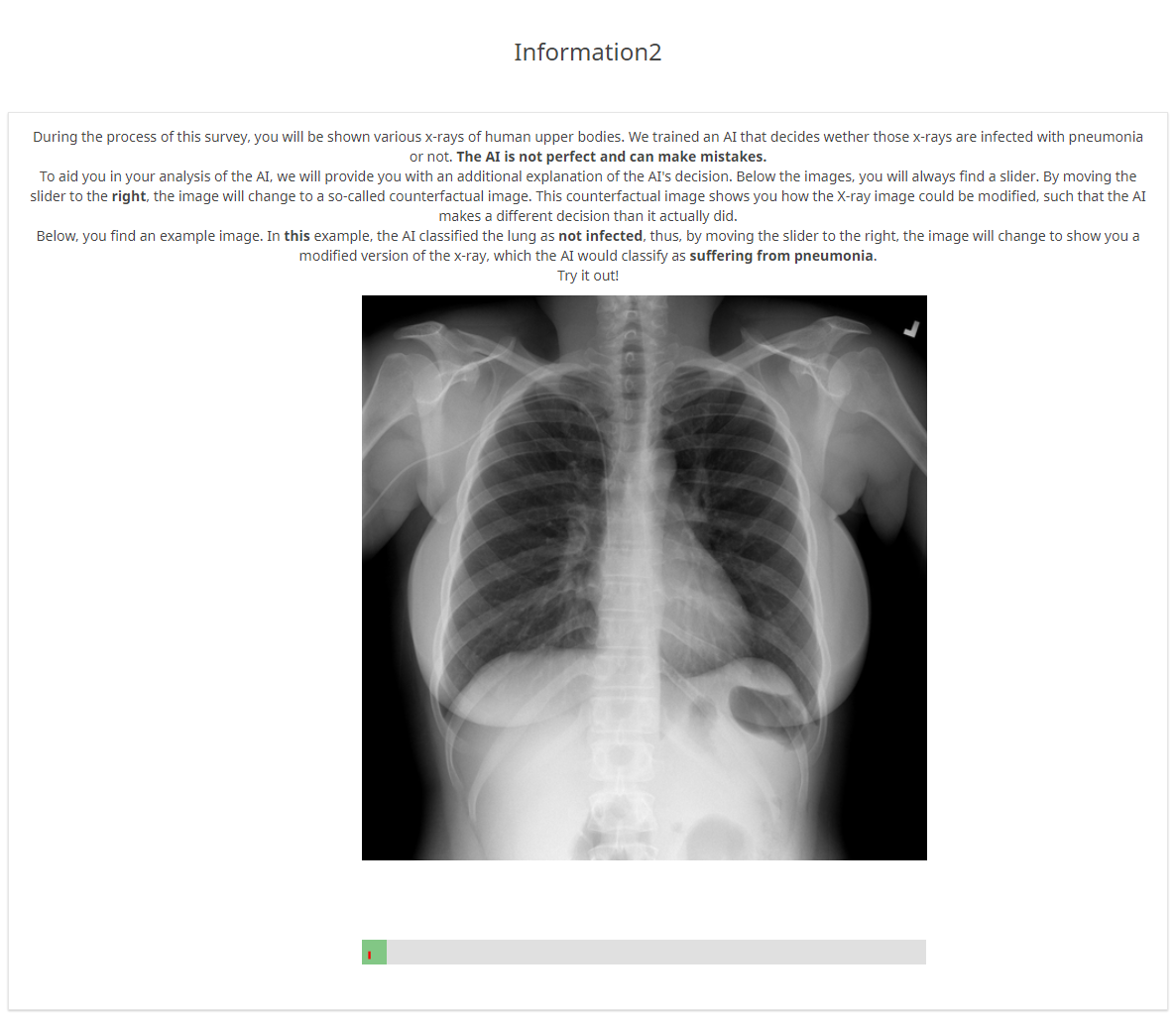}
  \includegraphics[width=1\linewidth]{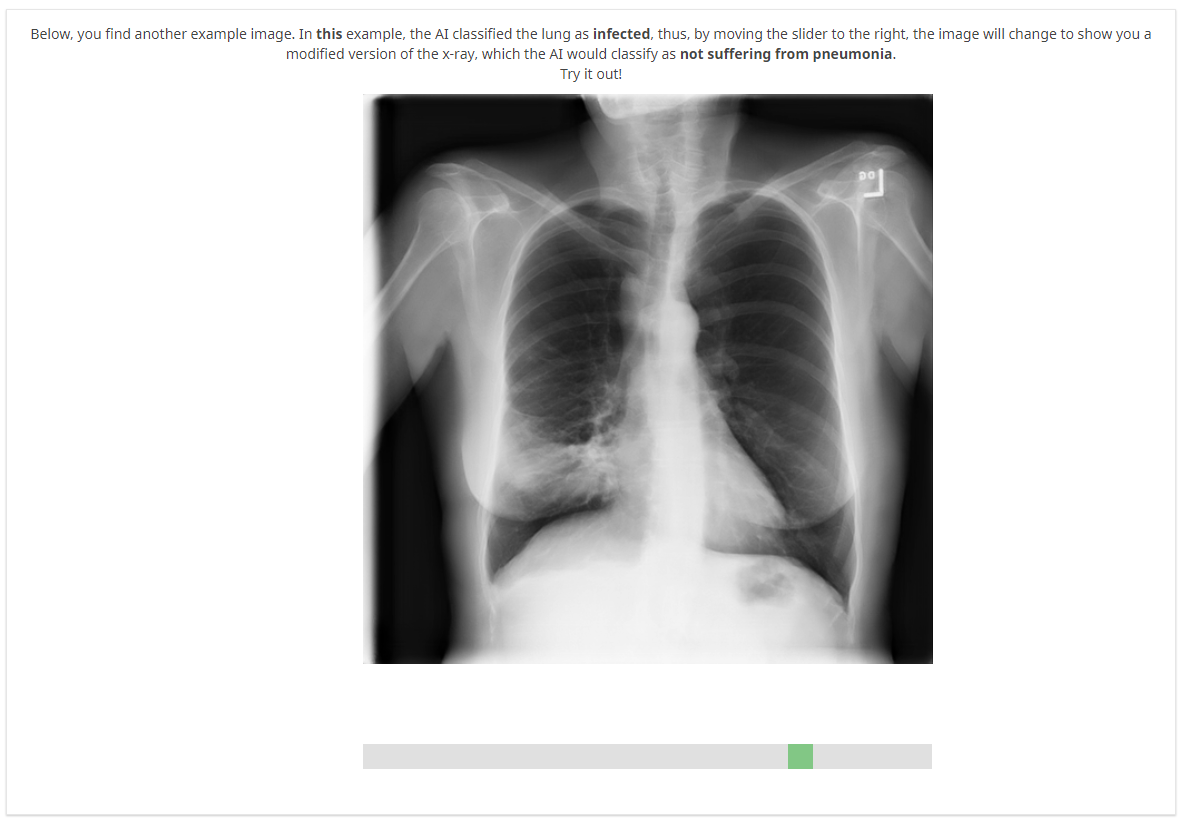}
\caption{Study Design}
\end{figure}

\begin{figure}[H]
 \centering
  \includegraphics[width=1\linewidth]{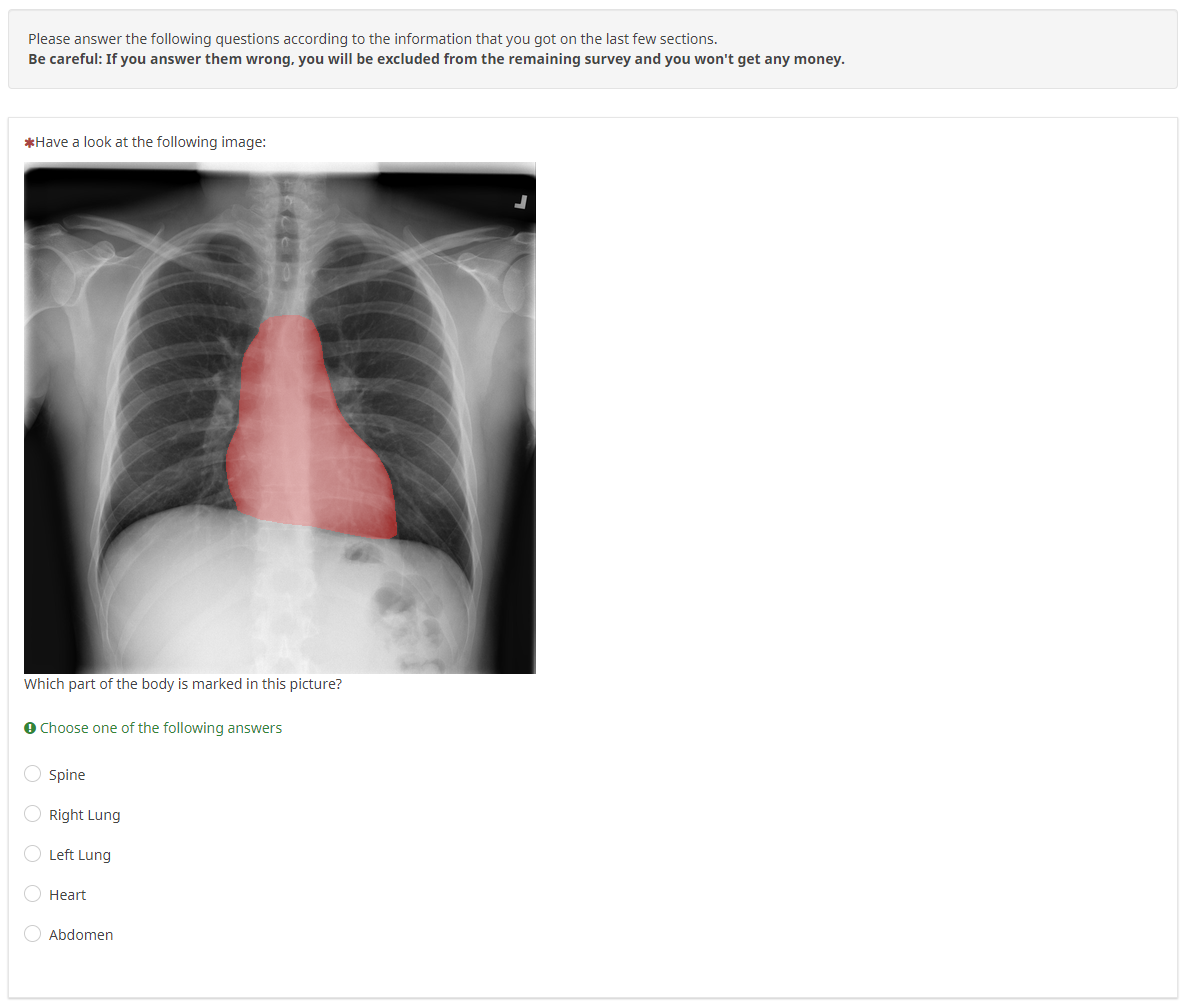}
  \includegraphics[width=1\linewidth]{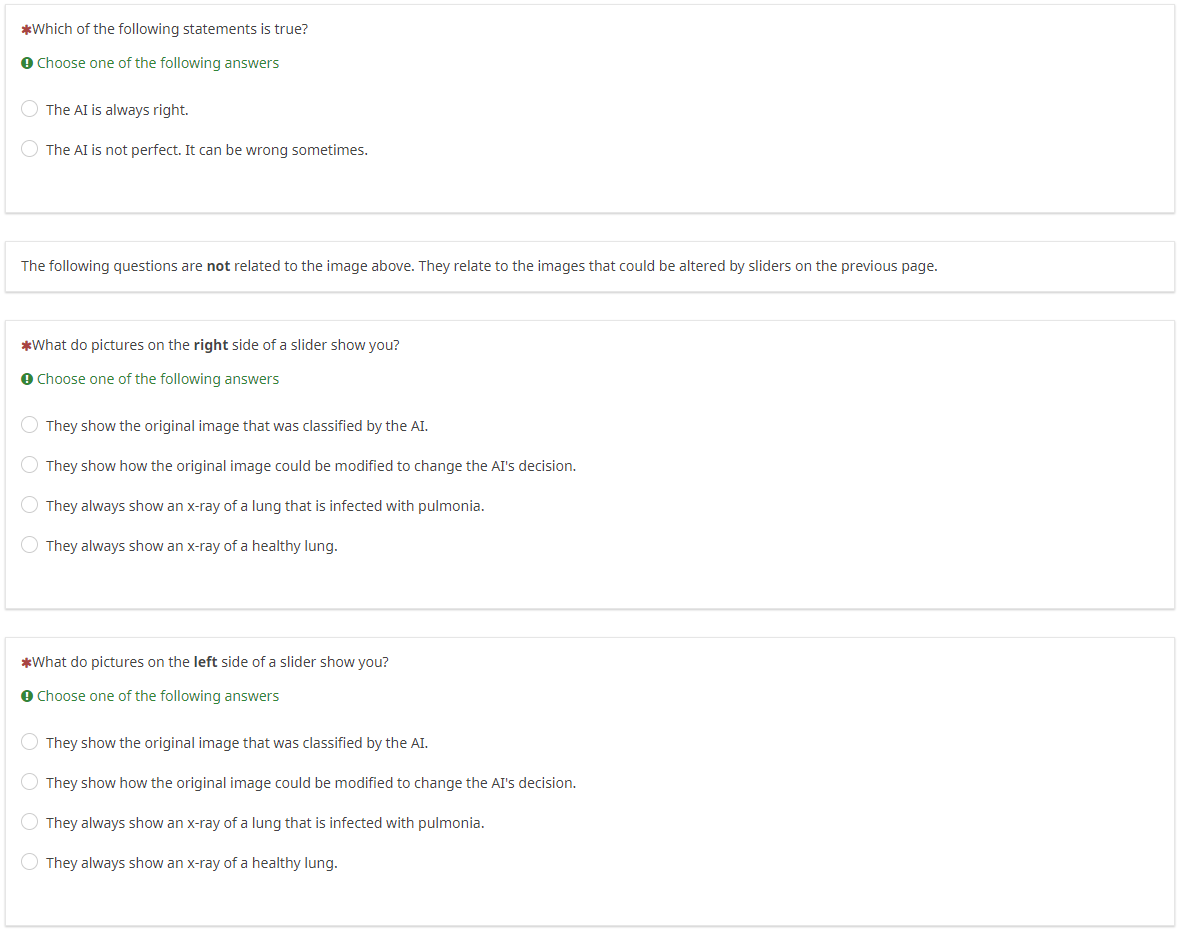}
\caption{Study Design}
\end{figure}

\begin{figure}[H]
 \centering
  \includegraphics[width=1\linewidth]{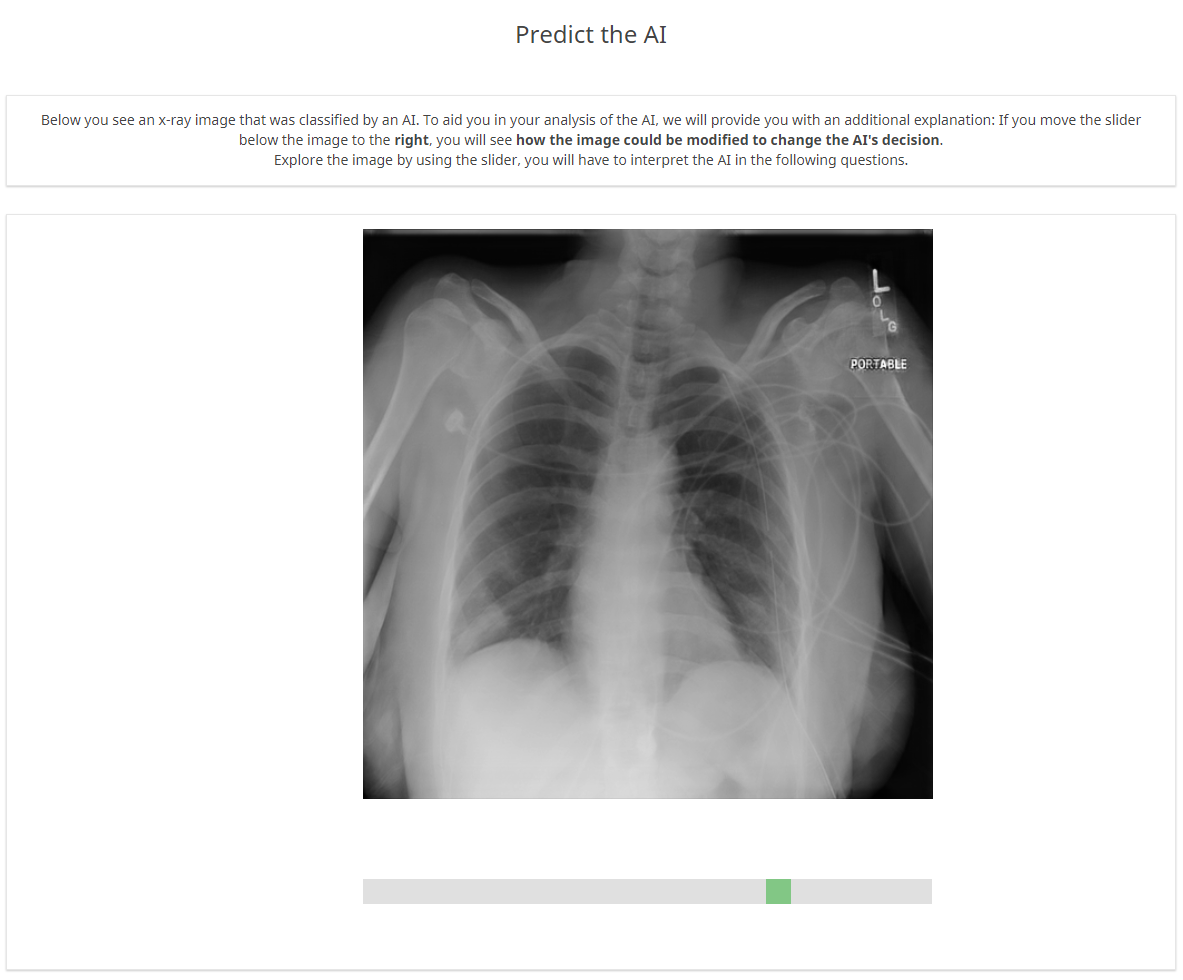}
  \includegraphics[width=1\linewidth]{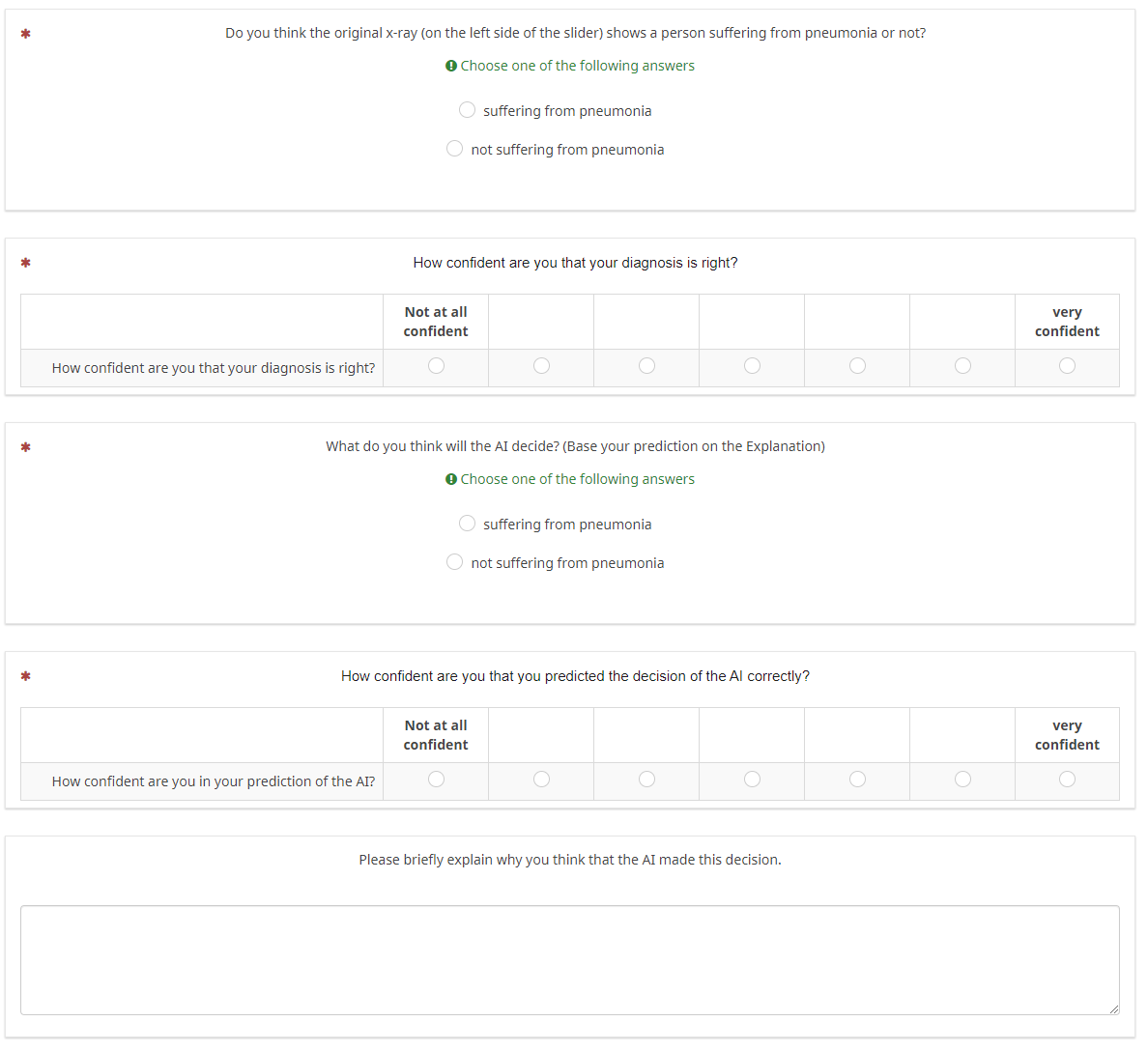}
\caption{Study Design}
\end{figure}

\begin{figure}[H]
 \centering
  \includegraphics[width=1\linewidth]{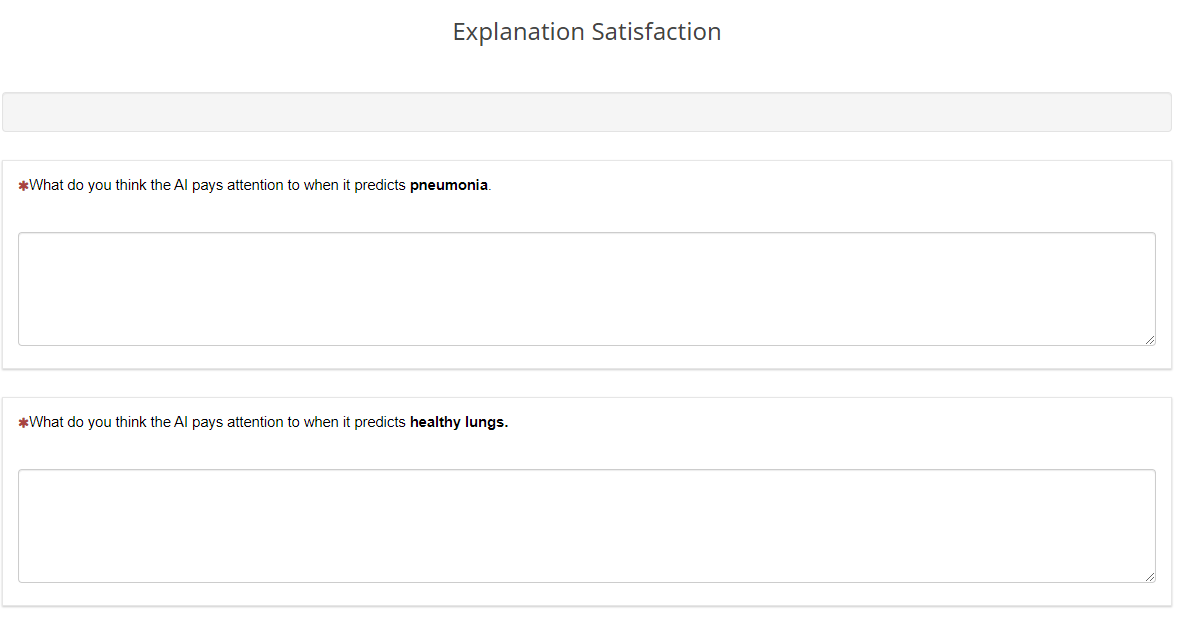}
  \includegraphics[width=1\linewidth]{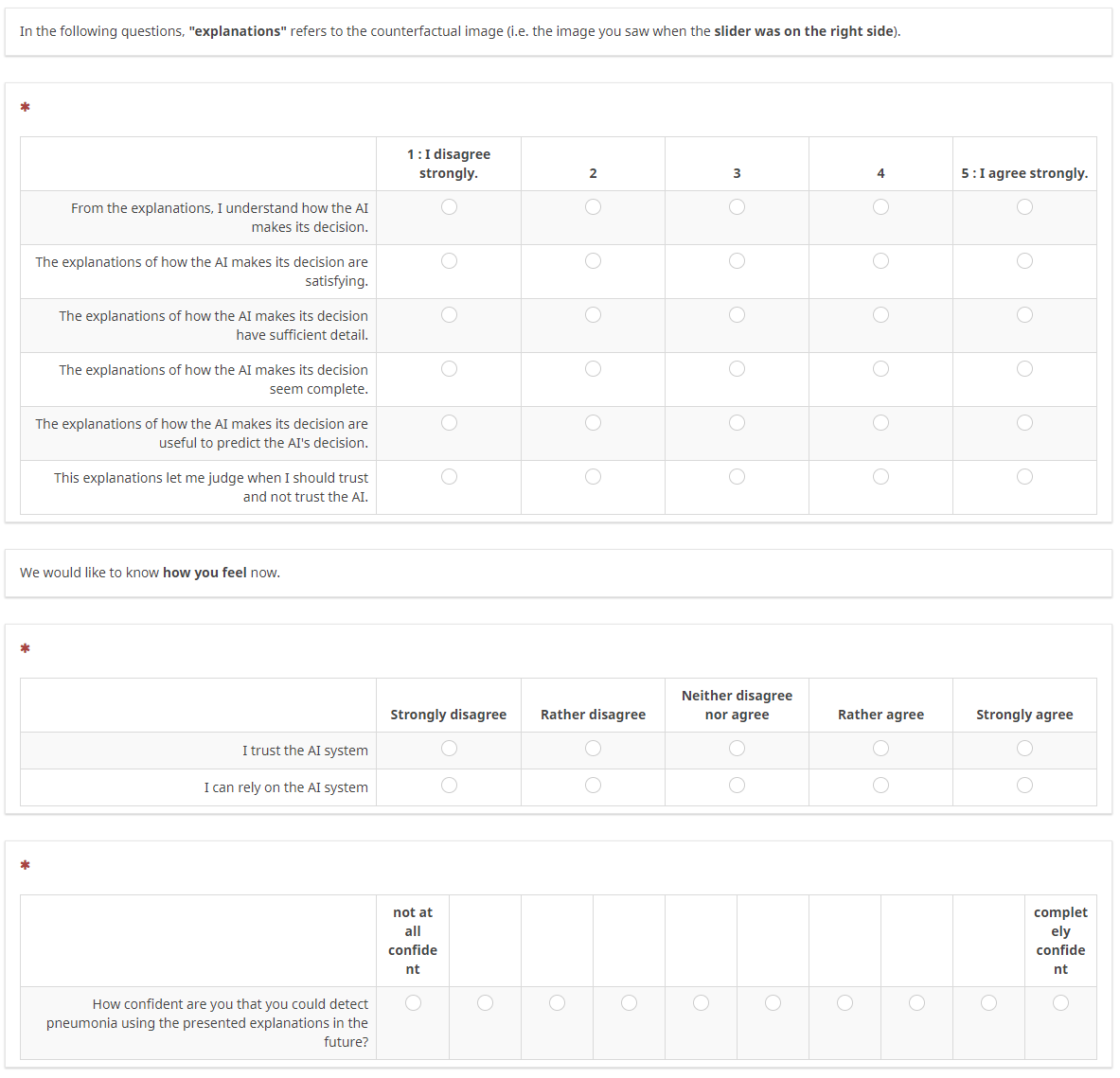}
\caption{Study Design}
\end{figure}

\begin{figure}[H]
 \centering
  \includegraphics[width=1\linewidth]{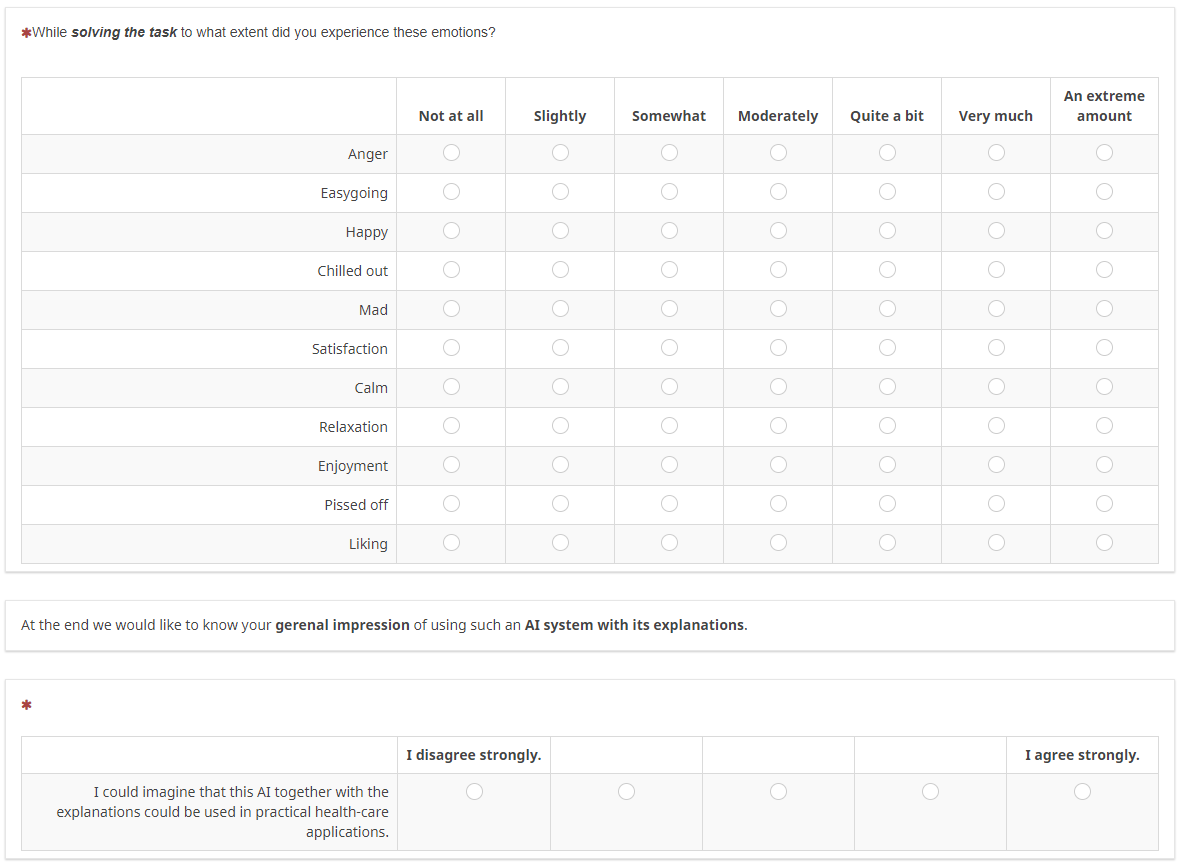}
  \includegraphics[width=1\linewidth]{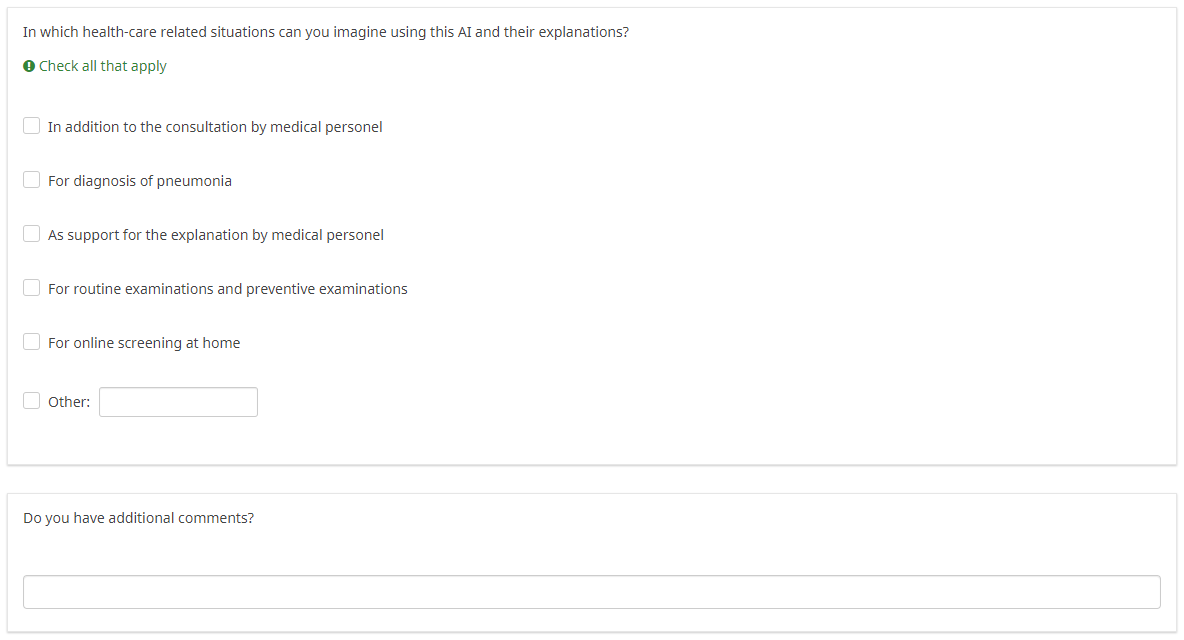}
\caption{Study Design}
\end{figure}

\end{document}